\documentclass{article}

\usepackage{microtype}
\usepackage{graphicx}
\usepackage{subfigure}

\usepackage{booktabs} 
\usepackage{hyperref}
\usepackage{enumitem}

\usepackage[accepted]{icml2025}

\usepackage{amsmath}
\usepackage{amssymb}
\usepackage{mathtools}
\usepackage{amsthm}
\usepackage[capitalize,noabbrev]{cleveref}

\theoremstyle{plain}

\theoremstyle{definition}

\theoremstyle{remark}

\usepackage[textsize=tiny]{todonotes}

\icmltitlerunning{Prototype-Guided Classification Sub-Task Decoupling Framework...}

\begin{document}

\twocolumn[
\icmltitle{Prototype-Guided Classification Sub-Task Decoupling Framework: \\Enhancing Generalization and Interpretability for Multivariate Time Series}

\icmlsetsymbol{equal}{*}

\begin{icmlauthorlist}
\icmlauthor{Xianhao Song}{aff1}
\icmlauthor{Yuang Zhang}{aff1}
\icmlauthor{Yuqi She}{aff1}
\icmlauthor{Liping Wang}{aff1}
\icmlauthor{Xuemin Lin}{aff2}
\end{icmlauthorlist}

\icmlaffiliation{aff1}{East China Normal University}
\icmlaffiliation{aff2}{Shanghai Jiao Tong University}
\icmlcorrespondingauthor{Xuemin Lin}{51275902123@stu.ecnu.edu.cn}

\icmlkeywords{Machine Learning}


\vskip 0.3 in
]



\printAffiliationsAndNotice{}  

\begin{abstract}
Time Series Classification (TSC) is a long-standing research problem that has gained increasing attention in recent years with the rapid growth of large-scale temporal data. Despite substantial progress enabled by deep learning, designing TSC models that are both accurate and interpretable remains a challenging task. Many existing approaches adopt a direct feature-to-label classification paradigm, by collapsing high-dimensional temporal embeddings into class logits via a single linear projection (often after global pooling), the paradigm conflates feature extraction and decision logic into an inseparable mapping.
 
To address these limitations, we propose PDFTime, a prototype-guided framework that reformulates time series classification as a multi-stage decision process. Instead of direct feature-to-label mapping, PDFTime leverages learned prototypes to approximate class-conditional feature distributions in the latent space, enabling progressive discrimination through classification sub-tasks of varying granularity. To our knowledge, PDFTime is the first framework to reformulate time series classification as a decoupled, multi-stage similarity-based reasoning process, breaking the long-standing paradigm of direct, black-box feature-to-label mapping. Extensive evaluations demonstrate that PDFTime achieves state-of-the-art (SOTA) performance across UEA and UCR benchmarks. Notably, it secures the top-$1$ accuracy on 80 out of 128 datasets in the UCR archive, significantly outperforming recent strong baselines in both consistency and generalization.
\end{abstract}
\section{Introduction}
Time Series Classification (TSC) plays a critical role in a wide range of applications, including healthcare~\cite{vrba2001signal,tang2023detecting}, human action recognition~\cite{shokoohi2017generalizing,amaral2022summertime}, audio signal processing~\cite{ruiz2021great}, Internet of Things~\cite{bakirtzis2022deep}, and semantic communication~\cite{zhao2023classification}. Unlike data modalities such as images or natural language, time series data consist of ordered numerical observations with strong temporal dependencies and complex variable interactions. Discriminative patterns may appear at different temporal scales and are often entangled across dimensions, making effective representation learning and reliable decision reasoning inherently challenging.

Recent advances in deep learning have substantially improved the performance of time series classification models~\cite{seto2015multivariate,schafer2017multivariate,li2023sleep,tang2020omni,moderntcn}. A large body of existing work adopts end-to-end classification architectures, where convolutional or Transformer-based backbones are trained to directly optimize downstream classification objectives. Representative methods such as ModernTCN~\cite{moderntcn} demonstrate that carefully designed architectures and inductive biases can achieve strong performance across diverse benchmarks. In parallel, another line of research explores large-scale and self-supervised pretraining for time series modeling, including methods such as TS-TCC~\cite{tstcc} and GPT-2-based sequence models~\cite{gpt2}, which aim to learn general-purpose temporal representations transferable across tasks and datasets. Together, end-to-end models and pretrained representation models constitute two complementary paradigms in modern TSC research.
\begin{figure}[htbp]
    \centering
    \includegraphics[width=1\linewidth]{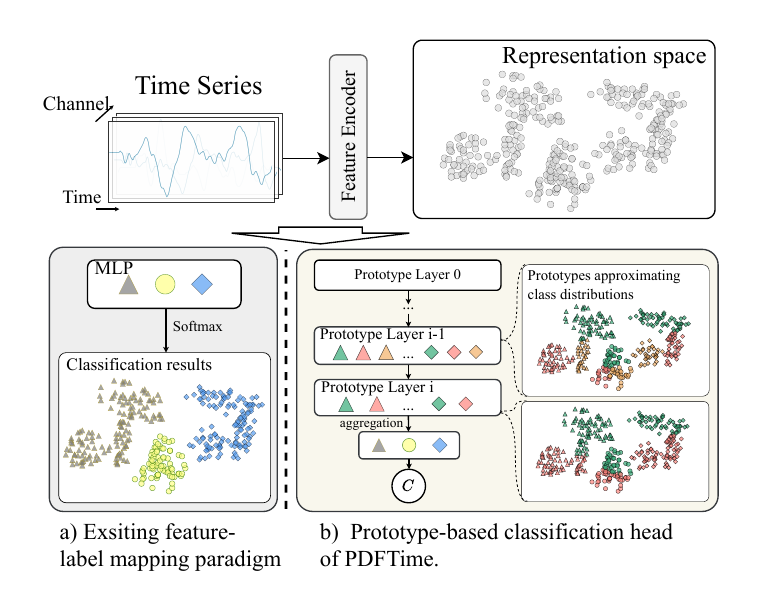}
    \vspace{-2mm}
    \caption{a) Existing methods map the feature space directly to class categories and use a softmax operation to predict the class with the highest probability as the final output.
b)\textbf{ PDFTime} decomposes this process into multiple classification stages, enabling the model to gradually solve simpler discrimination tasks and progressively refine them into the final classification result.}
    \label{fig:intro}
    \vspace{-5mm}
\end{figure} 

\textbf{Challenge.}
Despite these advances, generalization remains a central challenge for both paradigms. For pretrained time series foundation models, adaptation to downstream classification tasks is still limited: commonly used strategies such as global pooling or unified sequence representations tend to weaken the utilization of locally discriminative temporal patterns, while semantic discrepancies between pretraining data distributions and specific classification scenarios further complicate representation transfer. For end-to-end models trained from scratch, performance improvements are primarily driven by stronger feature extraction capabilities, which can be effective across multiple domains but do not always translate into robust generalization under heterogeneous temporal characteristics.

In addition to generalization, interpretability poses another persistent challenge in TSC. Most existing approaches rely on a direct feature-to-label mapping, such as fig~\ref{fig:intro}(a), where high-dimensional temporal representations are compressed into class logits and normalized through a Softmax operation. While this design is effective for optimizing classification accuracy, it tightly couples representation learning and decision making, obscuring intermediate reasoning processes and offering limited insight into how fine-grained temporal evidence contributes to class-level predictions. These limitations highlight the need for classification frameworks that better bridge temporal evidence and decision outcomes, while supporting more interpretable reasoning in time series classification.

To address these limitations, we propose \textbf{PDFTime}, a \textbf{P}rototype-Guided Classification Sub-Task \textbf{D}ecoupling \textbf{F}ramework for time series classification.
Instead of directly mapping extracted features to class logits, PDFTime formulates prediction as an explicit similarity aggregation process and decomposes classification into a sequence of progressively refined sub-tasks.
To illustrate this limitation and our solution, as illustrated in Fig~\ref{fig:intro} (b), while conventional methods collapse features into a single logit, PDFTime explicitly structures the latent space into class-conditional prototype clusters, facilitating a more granular and interpretable decision trajectory. Our code is available at \url{https://anonymous.4open.science/r/PDFTime-F2CA}.

Our \textbf{contributions} are summarized as follows:
\begin{itemize}[leftmargin=*]
    \vspace{-2mm}
    \item We propose \textbf{PDFTime}, a prototype-guided classification framework for multivariate time series classification that decouples temporal representation learning from the final decision-making process.
    \vspace{-2mm}
    \item We design a novel prototype-based classification head that performs structured, similarity-driven inference through a hierarchical organization of prototypes, enabling not only multi-granularity reasoning but also transparent attribution of predictions to specific prototype matches.
    \vspace{-2mm}
    \item Extensive evaluations across diverse benchmarks demonstrate the superior generalization capability of PDFTime. Notably, PDFTime achieves \textbf{state-of-the-art (SOTA) results on 80 out of 128 datasets} in the UCR archive, setting a new performance ceiling for the community and consistently outperforming recent competitive baselines.
\end{itemize}

\section{Related Work}

\textbf{Deep Learning for Time-Series Classification.}
Deep learning has become the dominant paradigm for time series classification (TSC) due to its strong ability to automatically extract discriminative temporal representations~\cite{test,tslanet}. Compared with traditional distance-based or feature-engineered methods, deep neural networks directly model raw temporal signals and capture complex non-linear dependencies.

Existing approaches are mainly categorized by architectural design. Convolutional neural networks (CNNs) exploit local receptive fields and hierarchical feature aggregation to model short-term patterns, leading to effective architectures such as InceptionTime and ModernTCN~\cite{moderntcn}. Transformer-based models, on the other hand, leverage self-attention to capture long-range dependencies and global temporal interactions, achieving competitive performance on both univariate and multivariate benchmarks~\cite{fedformer,patchTST}. In parallel, self-supervised learning methods, like TSTCC\cite{tstcc}, learn transferable temporal representations via contrastive objectives, reducing reliance on labeled data and improving robustness.

Despite these advances, most deep TSC methods—whether supervised or self-supervised—still adopt monolithic classification heads that tightly couple representation learning with discrimination, providing limited insight into the decision process and potentially hindering generalization under distribution shifts.

\textbf{Interpretable Time-Series Classification.}
Interpretability has long been a central concern in time series classification, motivating methods that associate predictions with meaningful temporal structures.

Recent work increasingly integrates interpretability into deep learning frameworks. Shapelet-based approaches embed discriminative subsequence discovery into neural architectures, allowing models to highlight informative temporal segments while maintaining strong representation learning capacity~\cite{qu2024cnn,le2024shapeformer,wen2025shedding}. Segment-based techniques further abstract time series into structured temporal components and achieve competitive performance across diverse datasets~\cite{moderntcn,medformer,wen2024abstracted}.

Another line of research explores interpretability through frequency-aware modeling. By incorporating frequency-domain representations or spectral priors, these methods relate predictions to global temporal dynamics or periodic structures~\cite{wang2022learning,woo2022cost,tslanet,fedformer}. In multivariate settings, channel-aware mechanisms have also been proposed to assess feature importance across dimensions~\cite{kim2024cafo}.

Overall, existing interpretable TSC methods primarily focus on explaining feature extraction or importance estimation. However, the mapping from learned representations to final class decisions is typically handled by implicit classification heads, limiting transparency at the decision-making stage.
\textbf{Prototype Learning for Time Series Analysis.}
Prototype learning represents each class by one or multiple prototypes in a latent space and performs classification via similarity-based reasoning (see Section~\ref{sec:3.2}). This paradigm enables structured and interpretable decision processes, as predictions are explicitly grounded in prototype-level comparisons.

Prototype-based learning has been extensively explored in computer vision. In dense prediction tasks, prototypes serve as class-level anchors that organize feature spaces and improve generalization across data distributions~\cite{zhou2022rethinking}. Methods such as ProMotion~\cite{lu2024promotion} and ProtoSEG~\cite{qin2023unified} show that prototype-driven reasoning enhances both robustness and interpretability.

In contrast, prototype learning remains relatively underexplored in time series analysis. Most existing TSC models rely on implicit feature-to-label mappings, offering limited insight into how temporal representations correspond to semantic categories. While recent studies such as TEST~\cite{test} and TimeDP~\cite{timedp} incorporate prototype-related concepts, they mainly focus on representation alignment or generative modeling rather than explicit prototype-guided classification.

As a result, prototype-based decision mechanisms that decouple representation learning from classification are still largely absent in TSC. Given the sequential and hierarchical nature of time series data, this gap motivates our work, which introduces a prototype-guided framework for transparent and structured decision making in TSC.

\section{Preliminary}

\subsection{Problem Definition}
Multivariate time series classification is defined as the task of assigning a class label to each multivariate time series.
Formally, given a dataset
$D = \{(S_1, y_1), \ldots, (S_n, y_n)\}$,
each sample $S_i$ is associated with a class label
$y_i \in \{1, 2, \ldots, C\}$, where $C$ denotes the number of classes.

Following common preprocessing protocols in multivariate time series classification,
each sample $S_i$ is resampled and represented as a fixed-length multivariate sequence:
\begin{equation}
\mathbf{X}_i \in \mathbb{R}^{V \times L},
\end{equation}
where $V$ denotes the number of variables and $L$ is the unified temporal length.
This representation serves as the input to the proposed embedding network.
The objective is to learn a mapping from $\mathbf{X}_i$ to its corresponding class label
$y_i$.

\subsection{Prototype Learning}
\label{sec:3.2}
Let $E_\theta(\cdot)$ be an embedding network (e.g., a CNN or Transformer) such that $\mathbf{z}_i = E_\theta(\mathbf{X}_i) \in \mathbb{R}^d$ represents the latent feature of sample $i$. A \textbf{prototype} is a representative point in the feature space that summarizes characteristic patterns of a class.
Given an embedding space $\mathbb{R}^d$ and a set of classes
$Y = \{1, 2, \ldots, C\}$, each class $c \in Y$ can be associated with one or more
prototypes:
\begin{equation}
    \mathbf{p}^c_k \in \mathbb{R}^d, \quad k = 1, \ldots, K_c,
\end{equation}
where $K_c$ denotes the number of prototypes for class $c$.
Each prototype $\mathbf{p}^c_k$ represents a characteristic feature pattern of the
corresponding class.

Given a sample embedding $\mathbf{z} \in \mathbb{R}^d$, similarity to a prototype is
measured using a function
\begin{equation}
    \mathrm{sim}(\mathbf{z}, \mathbf{p}_k^c) = f(\mathbf{z}, \mathbf{p}_k^c),
\end{equation}
where $f(\cdot)$ is cosine similarity.
Unlike traditional softmax-based classifiers that partition the feature space with hyperplanes, prototype learning performs classification by comparing similarities in the latent space. This allows the model to ground its decisions in explicit representative patterns, facilitating the sub-task decoupling proposed in this work.
\section{Main Framework} 
\subsection{Model Overview}
\begin{figure}
    \centering
    \includegraphics[width=1\linewidth]{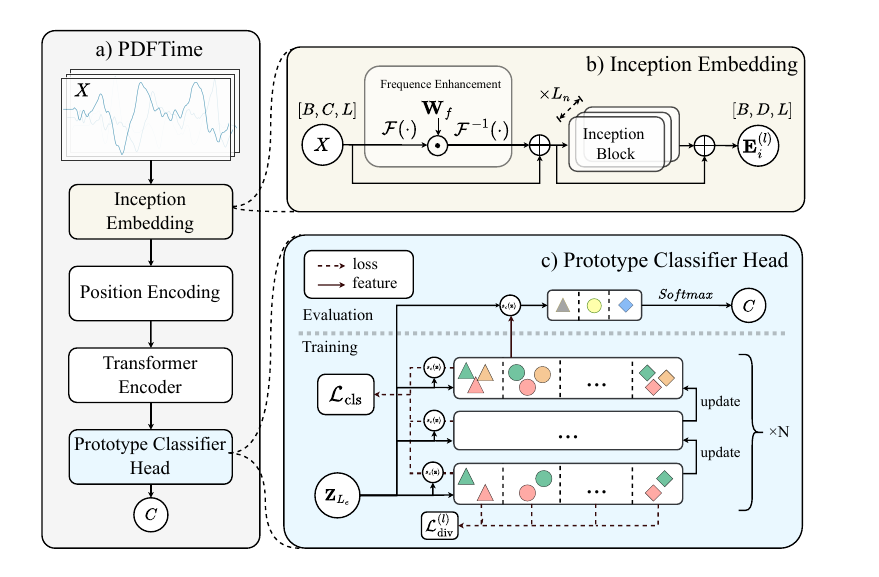}
    \vspace{-2mm}
    \caption{(a) Overall architecture of PDFTime. The model employs an inception-style embedding module (b) to enhance local feature extraction and utilizes a prototype-based classification head (c) to capture class-specific patterns at different levels of granularity.}
    \label{fig:framework}
    \vspace{-5mm}
\end{figure}

We now provide an overview of the proposed \textbf{PDFTime} framework.
As illustrated in Figure~\ref{fig:framework}, \textbf{PDFTime} is a prototype-guided classification framework designed to decouple temporal representation learning from decision making in time series classification.
The motivation for PDFTime is to introduce an explicit prototype-based inference stage, allowing structured reasoning over temporal representations and supporting interpretable and generalizable decision making.
The overall architecture consists of three functional components: an embedding module for local pattern extraction, a Transformer encoder for temporal modeling, and a prototype-based classification head for decision making.

\textbf{Inception Embedding Stage.}
The purpose of this stage is to enhance the expressive capacity of temporal representations at early stages.
To achieve this, PDFTime adopts an inception-style multi-scale convolutional embedding.
A frequency-weighting mechanism is incorporated to emphasize periodic structures, and the resulting embeddings are forwarded to the Transformer encoder.
Detailed descriptions of the embedding design are provided in Section~\ref{sec:4.2}.

\textbf{Transformer Encoder Stage.}
The primary purpose of this stage is to demonstrate that the effectiveness of PDFTime stems from the proposed classification framework rather than architectural sophistication in the backbone.
Accordingly, PDFTime deliberately adopts a standard $N$-layer vanilla Transformer encoder, without incorporating recent task-specific modifications or advanced architectural improvements.
By relying on this minimal and widely used design, we ensure that performance gains can be attributed to the prototype-guided classification mechanism itself.
The encoder applies self-attention with positional encoding to preserve temporal order, and further architectural details are presented in Section~\ref{sec:4.3}.

\textbf{Prototype-based Classification Stage.}
The objective of this stage is twofold: to provide interpretable semantic anchors for similarity-based reasoning, and to encourage the model to capture multi-granularity class-specific temporal patterns.
To this end, class-aware prototypes are organized hierarchically and used to perform similarity-based inference over the encoded representations.
The detailed formulation and update mechanism of the prototype classification head are described in Section~\ref{sec:4.4}.

\vspace{-2mm}
\subsection{Inception embedding}\label{sec:4.2}
Recent studies \cite{middlehurst2024bake} have demonstrated that CNN-based architectures exhibit strong empirical performance on time series classification (TSC) tasks.
Motivated by this observation, we adopt a multi-scale convolutional encoder inspired by InceptionTime \cite{inceptiontime} as the front-end embedding module.
By employing convolutional kernels of varying temporal sizes, the encoder is able to capture both short-term local fluctuations and longer-term temporal motifs.
Such multi-resolution representations provide a crucial foundation for prototype learning, enabling prototypes to distinguish class-specific patterns at different levels of granularity.

To mitigate the sensitivity of prototype learning to noisy and irregular signals, we further enhance the embedded features in the frequency domain.
Given a batch of intermediate feature representations $\mathbf{X} \in \mathbb{R}^{B \times V \times L}$, we first apply a discrete Fourier transform (DFT) along the temporal dimension:
\begin{equation}
\hat{\mathbf{X}}_i = \mathcal{F}(\mathbf{X}_i)
\end{equation}
where $\mathcal{F}(\cdot)$ denotes the Fourier transform.
A learnable frequency weighting mask is then applied to emphasize informative periodic components while suppressing noise:
\begin{equation}
\tilde{\mathbf{X}}_i = \mathcal{F}^{-1}\left( \mathbf{W}_f \odot \hat{\mathbf{X}}_i \right)
\end{equation}
where $\mathbf{W}_f$ is a learnable frequency mask and $\odot$ denotes element-wise multiplication.

The frequency-enhanced features $\tilde{\mathbf{X}}_i$ are subsequently processed by a stack of $L_I$ Inception encoding layers with residual connections.

Each Inception layer aggregates temporal patterns at multiple scales through parallel transformation branches.

Formally, let $e = \mathbf{E}_i^{(l-1)} \in \mathbb{R}^{D \times L}$ denote the input feature map of the $l$-th ($l \in \{1,\ldots,L_n \}$) Inception layer.
An Inception layer computes the output feature map through multi-branch parallel operations as follows:
\begin{equation}
\text{Inception}(e)
= \phi \Big(
\text{Concat}\big(
f_1(e),
f_2(e),
\ldots,
f_p(e)
\big)
\Big)
\end{equation}

where $\{f_k(\cdot)\}_{k=1}^{K}$ denote parallel temporal feature extraction branches operating at different receptive scales, 
$\text{Concat}(\cdot)$ represents channel-wise concatenation, and $\phi(\cdot)$ is a linear projection that aligns the concatenated features to a unified embedding space.

The output of the $l$-th Inception layer is then obtained via a residual formulation:
\begin{equation}
\mathbf{E}_i^{(l)} = \text{Inception}\left(\mathbf{E}_i^{(l-1)}\right) + \mathbf{E}_i^{(l-1)}
\end{equation}
where $\mathbf{E}_i^{(0)} = \tilde{\mathbf{X}}_i$.

\subsection{Transformer Encoder}\label{sec:4.3}

Following the feature encoding stage, the embedded representations are processed
by a standard Transformer encoder to model long-range temporal dependencies.
Let $\mathbf{E} \in \mathbb{R}^{B \times L \times D}$ denote the embedded feature
sequence produced by the embedding module, where $B$ is the batch size, $L$ the
sequence length, and $D$ the feature dimension.

To preserve temporal order, positional encoding is added to the embedded features:
\begin{equation}
\mathbf{Z}_0 = \mathbf{E} + \mathbf{P}
\end{equation}
where $\mathbf{P} \in \mathbb{R}^{L \times D}$ denotes the positional encoding,
which is broadcast along the batch dimension.
We follow the original sinusoidal positional encoding formulation introduced
in the Transformer architecture\cite{2017Attention}.

The position-enhanced representations are then passed through an $L_e$-layer
Transformer encoder.
Each layer consists of a multi-head self-attention (MHSA) module and a
position-wise feed-forward network (FFN), together with residual connections
and layer normalization.
Formally, the $l$-th ($l \in \{1,\ldots,L_e \}$) encoder layer is defined as:
\begin{align}
\mathbf{Z}_{l}' &= \mathrm{MHSA}(\mathrm{LN}(\mathbf{Z}_{l-1})) + \mathbf{Z}_{l-1} \\
\mathbf{Z}_{l}  &= \mathrm{FFN}(\mathrm{LN}(\mathbf{Z}_{l}')) + \mathbf{Z}_{l}'
\end{align}

After stacking $L_e$ layers, the final temporal representation
$\mathbf{Z}_{L_e} \in \mathbb{R}^{B \times L \times D}$ is obtained and forwarded
to the prototype-based classification module.

\subsection{Prototype-based Classification Module.}\label{sec:4.4}

\textbf{Design Principle.}
We introduce a decoupled prototype-based classification head, where each class is
represented by a set of prototypes that are not updated via backpropagation.
Unlike conventional learnable prototypes, this design prevents gradient-driven
sensitivity to initialization and avoids early overfitting to prototype parameters.
As a result, the backbone network is encouraged to learn more robust and transferable
embeddings, while the classification head focuses on similarity-based reasoning.

\textbf{Prototype-based Classification.}
Let $\{\mathbf{p}_{k}^{c}\}_{k=1}^{K_c}$ denote the set of prototypes associated with
class $c$, where each prototype $\mathbf{p}_{k}^{c} \in \mathbb{R}^D$ lies in the same
embedding space as the feature representation $\mathbf{z}$.
To capture intra-class variability at multiple granularities, we adopt a multi-level
prototype design, where each level contains a different number of prototypes per class.
The specific configurations of prototype layers and cardinalities are provided in
Appendix~\ref{app:prototype}.

At initialization, prototypes are constructed in a class-wise manner.
For each class, prototype vectors are randomly sampled and orthogonalized via QR decomposition to promote diversity and reduce redundancy.
All prototypes are then normalized to lie on a hypersphere with a fixed radius.
Throughout training, prototypes are treated as non-parametric memory vectors and are
not updated via gradient descent.

The similarity between an input feature $\mathbf{z}$ and a prototype is computed using
cosine similarity.
Class-level similarity scores are obtained by aggregating similarities over all
prototypes of the same class using a Log-Sum-Exp formulation:
\begin{equation}
s_c(\mathbf{z}) = \log \sum_{k=1}^{K_c}
\exp\left( \frac{\mathrm{sim}(\mathbf{z}, \mathbf{p}_k^c)}{T} \right)
\end{equation}
where $T$ is a temperature parameter controlling the sharpness of the aggregation.
Final predictions are produced by applying a softmax over the class-level scores
$\{s_c(\mathbf{z})\}$.

This design enables each class to be represented by multiple, diverse prototypes at
different granularities, allowing the model to effectively capture intra-class
variability and multi-modal temporal patterns commonly observed in real-world time series data.

\textbf{Prototype Update Strategy.}
Since prototypes are not optimized via backpropagation, we adopt a statistics-driven update mechanism to adapt prototypes during training. Specifically, prototypes are updated using an exponential moving average (EMA) scheme based on feature embeddings of training samples.
For a prototype $\mathbf{p}^c_k$ associated with class $c$, its update rule is defined as:
\begin{equation}
    \tilde{\mathbf{p}}_k^c =
    \frac{\sum_{i:y_i=c} q_{ik}\,\mathbf{z}_i}
         {\sum_{i:y_i=c} q_{ik}}, \quad
    q_{ik} = \mathrm{softmax}\left(
    \mathrm{sim}(\mathbf{z}_i,\mathbf{p}_k^c)
    \right)
\end{equation}
\begin{equation}
\mathbf{p}_k^{c,(t)} =
\gamma(t)\,\mathbf{p}_k^{c,(t-1)} +
(1-\gamma(t))\,\tilde{\mathbf{p}}_k^c
\end{equation}

where ${\mathbf{z}_i}$ are feature representations of training samples belonging to class $c$ within the current mini-batch,where $i \in \mathcal{B}_c$ $\mathcal{B}_c$ is the sample index set belonging to the category $c$ in the current batch and $(t)$ denotes the training step and $\gamma \in (0,1]$ is a coefficient that controls the update speed.
Rather than using a fixed $\gamma$, we employ a dynamic schedule that adapts $\gamma$ over training steps to balance stability and adaptability. The scheduling logic is summarized in Algorithm~\ref{alg:gamma_schedule}. In our implementation, we use $T_{\text{warm}} = 3$, $T_{\text{active}} = 10$, $\gamma_a = 0.99$, $\gamma_b = 0.999$, and $\tau = 30$. We observe that PDFTime is not highly sensitive to moderate variations of $\gamma_a$ and $\gamma_b$, and we report an ablation study in Appendix ~\ref{app:prototype}.

\begin{algorithm}[t]
\caption{Dynamic $\gamma$ Scheduling (Training Phase)}
\label{alg:gamma_schedule}
\begin{algorithmic}[1]
\REQUIRE Current step $t$, warm-up steps $T_{\text{warm}}$, active steps $T_{\text{active}}$, lower momentum $\gamma_a$, upper momentum $\gamma_b$, time constant $\tau$
\ENSURE Momentum coefficient $\gamma(t)$

\IF{$t < T_{\text{warm}}$}
    \STATE $\gamma \gets 1.0$ \hfill \textit{// Warm-up: freeze prototypes}
\ELSIF{$t < T_{\text{warm}} + T_{\text{active}}$}
    \STATE $\alpha \gets \dfrac{t - T_{\text{warm}}}{T_{\text{active}}}$
    \STATE $\gamma \gets 1.0 - (1.0 - \gamma_a) \cdot \alpha$ \hfill \textit{// Linear decay to $\gamma_a$}
\ELSE
    \STATE $\Delta t \gets t - T_{\text{warm}} - T_{\text{active}}$
    \STATE $\gamma \gets \gamma_a + (\gamma_b - \gamma_a) \cdot \left(1 - e^{-\Delta t / \tau}\right)$ \hfill \textit{// Exponential approach to $\gamma_b$}
\ENDIF

\end{algorithmic}
\end{algorithm}
This three-phase design ensures that:
(i) prototypes remain fixed during early training to stabilize feature learning;
(ii) they become responsive during the adaptation phase to capture emerging cluster structures; and
(iii) they gradually lock into stable centroids in later stages by increasing the EMA memory length.

At inference time, all prototypes are fixed and no further updates are performed.
Although multiple prototype levels are maintained during training, only the final (highest-level) prototypes are used for similarity-based inference and prediction. This strategy enables smooth and stable decision making based on consolidated class-level representations.

\textbf{Prototype Diversity Regularization.}
Formally, let $\{\mathbf{p}^c_1,\dots,\mathbf{p}^c_K\}$ denote the $K$ prototypes associated with class $c$ at a given prototype level. We construct a class-wise prototype matrix $\mathbf{P}^c \in \mathbb{R}^{K \times D}$ and compute the intra-class similarity matrix
\begin{equation}
\mathbf{S}^c = \mathbf{P}^c (\mathbf{P}^c)^\top \in \mathbb{R}^{K \times K}
\end{equation}

To encourage diversity among prototypes of the same class, we enforce $\mathbf{S}^c$ to approximate an identity matrix, such that prototypes remain distinct while preserving self-similarity. The intra-class prototype diversity loss is defined as
\begin{equation}
\mathcal{L}_{\text{div}} = \frac{1}{C} \sum_{c=1}^{C} \left\| \mathbf{S}^c - \mathbf{I} \right\|_2^2 
\end{equation}
where $C$ denotes the number of classes and $\mathbf{I}$ is the identity matrix of size $K \times K$.

During training, the overall objective at each prototype level is computed by combining the classification loss and the diversity regularization:
\begin{equation}
\mathcal{L} = \sum_{l} (w_l\,\mathcal{L}^{(l)}_{\text{cls}} + \lambda\,\mathcal{L}^{(l)}_{\text{div}})
\end{equation}
where $w_l$ is the predefined weight for the classification loss at level $l$, and $\lambda=0.01$ controls the contribution of the prototype diversity regularization.

\textbf{Interpretability.} PDFTime offers intrinsic transparency via similarity-based attribution. Predictions are grounded in proximity to representative patterns $\{\mathbf{p}^c_k\}$ rather than opaque decision boundaries. This structured latent space provides a coarse-to-fine reasoning path: input $\mathbf{z}$ is ``explained" by its match with specific class-representative motifs, facilitating verifiable decision-making without post-hoc tools.

\vspace{-2mm}
\section{Experiments}
We focus on evaluating the generalization ability of the proposed framework across diverse datasets rather than optimizing performance on a single. Following~\cite{softshape,rocket}, we evaluate model performance using Top-1 count(only UCR datasets), average classification accuracy, and average rank across datasets.Top-1 count measures the number of datasets on which a method achieves the highest classification accuracy among all compared approaches.

\vspace{-2mm}
\subsection{Experimental Setup}

\begin{table*}[htbp]
\centering
\caption{Comparison of classification accuracy on 10 multivariate time series datasets in UEA. Best results are highlighted in  \textcolor{red}{red}, and second-best results in \textcolor{blue}{blue}.}
\resizebox{\textwidth}{!}{
\begin{tabular}{lcccccccccc}
\toprule
\textbf{Datasets} &
\textbf{Fedformer} &
\textbf{patchTST} &
\textbf{TSLANET} &
\textbf{ModernTCN} &
\textbf{FIC-TSC} &
\textbf{TimesNet} &
\textbf{NST} &
\textbf{GPT2} &
\textbf{ST-MEM} &
\textbf{PDFTime(Ours)} \\
\midrule
EthanolConcentration & 0.361 & 0.350 & 0.304 & 0.363 & \textcolor{red}{0.392} & 0.357 & 0.262 & 0.342 & 0.323 & \textcolor{blue}{0.369} \\
FaceDetection        & 0.671 & 0.656 & 0.668 & \textcolor{red}{0.708} & 0.684 & 0.686 & 0.673 & 0.692 & 0.644 & \textcolor{blue}{0.698} \\
Handwriting          & 0.212 & 0.188 & 0.579 & 0.306 & \textcolor{blue}{0.616} & 0.321 & 0.374 & 0.327 & 0.184 & \textcolor{red}{0.713} \\
Heartbeat            & 0.746 & 0.727 & 0.776 & 0.772 & \textcolor{red}{0.810} & 0.780 & 0.790 & 0.772 & 0.746 & \textcolor{blue}{0.800} \\
JapaneseVowels       & 0.954 & 0.954 & \textcolor{red}{0.992} & 0.898 & \textcolor{blue}{0.991} & 0.984 & 0.981 & 0.986 & 0.935 & 0.989 \\
PEMS-SF              & 0.827 & 0.832 & 0.618 & 0.838 & 0.792 & \textcolor{red}{0.896} & 0.855 & 0.879 & 0.770 & \textcolor{blue}{0.878} \\
SelfRegulationSCP1   & 0.573 & 0.812 & 0.918 & \textcolor{red}{0.934} & 0.901 & \textcolor{blue}{0.918} & 0.915 & \textcolor{blue}{0.932} & 0.877 & 0.887 \\
SelfRegulationSCP2   & 0.517 & 0.511 & 0.323 & \textcolor{red}{0.617} & \textcolor{blue}{0.594} & 0.572 & 0.572 & \textcolor{blue}{0.594} & 0.561 & 0.572 \\
SpokenArabicDigits   & 0.988 & 0.980 & \textcolor{blue}{0.999} & 0.987 & \textcolor{blue}{0.999} & 0.990 & 0.980 & 0.992 & 0.970 & \textcolor{red}{1.000} \\
UWaveGestureLibrary  & 0.703 & 0.856 & 0.903 & 0.867 & \textcolor{blue}{0.902} & 0.853 & 0.856 & 0.881 & 0.694 & \textcolor{red}{0.921} \\
\midrule
\textbf{Average Accuracy} &
0.655 & 0.687 & 0.708 & 0.729 & \textcolor{blue}{0.769} &
0.736 & 0.726 & 0.740 & 0.670 & \textcolor{red}{0.783} \\
\textbf{Average Rank} &
7.40 & 8.00 & 5.30 & 4.60 & \textcolor{blue}{3.30} &
4.60 & 5.90 & 4.10 & 9.00 & \textcolor{red}{2.80} \\
\bottomrule
\end{tabular}
}
\label{tab:uea_results}
\vspace{-5mm}
\end{table*}

\textbf{Benchmarks.}
 To evaluate the performance of different methods for time series classification (TSC), we conduct experiments on the UCR~\cite{ucr} and UEA~\cite{uea} time series archives, which are widely recognized benchmarks for univariate and multivariate TSC, respectively.

\textbf{Baselines.}
Following~\cite{moderntcn,timesnet,timemil}, We compare with the state-of-the-art methods of recent multivariate time series analysis on 10 selected UEA datasets. Including FedFormer~\cite{fedformer}, PatchTST~\cite{patchTST}, TSLANet~\cite{tslanet}, GPT2~\cite{gpt2}, NST~\cite{nst}, FIC-TSC~\cite{fic-tsc}, TEST~\cite{test}, ST-MEM~\cite{stmem}, and ModernTCN~\cite{moderntcn}.

For the UCR benchmarks, we compare our method with a broad range of strong TSC baselines, including TEST~\cite{test}, PatchTST~\cite{patchTST}, TSLANet~\cite{tslanet}, SARC~\cite{SARC}, SoftShape~\cite{softshape}, TimesNet~\cite{timesnet}, ROCKET~\cite{rocket}, and InceptionTime~\cite{inceptiontime}.

\textbf{Implementation details.}
For all baseline methods, we report the benchmark results as published in their original papers and follow the same experimental settings. Additional implementation details are provided in Appendix~\ref{app:fair}.

\vspace{-2mm}
\subsection{Experiment Results}
\textbf{Result on UEA benchmark.}
Table~\ref{tab:uea_results} summarizes the results on the UEA benchmark. PDFTime achieves an average accuracy of 78.3\%, outperforming the strongest baseline, FIC-TSC, by 1.4\%. This substantial margin underscores the effectiveness of our decoupled framework in handling complex multivariate interactions.

\textbf{Result on UCR benchmark.}
\begin{table}[t]
\centering
\caption{Performance comparison on the 128 UCR datasets. 
Top-1 count indicates the number of datasets on which a method achieves the best accuracy. Best results are highlighted in \textcolor{red}{red}, and second-best results in \textcolor{blue}{blue}.}
\label{tab:ucr_main_results}
\begin{tabular}{lccc}
\toprule
\textbf{Method} & \textbf{Top-1 num} & \textbf{Avg. Acc} & \textbf{Avg. Rank} \\
\midrule
TEST          & 22.0 & 0.8212 & 5.8594 \\
PatchTST      & 11.0 & 0.8255 & 5.6641 \\
TSLANet       & 16.0 & 0.9210 & 3.4688 \\
SARC          & 10.0 & 0.8508 & 5.6094 \\
SoftShape     & \textcolor{blue}{26.0} & \textcolor{blue}{0.9326} & \textcolor{blue}{2.9609} \\
TimesNet      & 6.0  & 0.8367 & 5.8516 \\
ROCKET        & 11.0 & 0.8450 & 5.6250 \\
InceptionTime & 7.0  & 0.8352 & 6.1875 \\
\midrule
\textbf{PDFTime(Ours)} & \textcolor{red}{80.0} & \textcolor{red}{0.9461} & \textcolor{red}{2.5938} \\
\bottomrule
\end{tabular}
\vspace{-3mm}
\end{table}
The results on 128 UCR datasets are reported in Table~\ref{tab:ucr_main_results}. PDFTime achieves an average accuracy of 94.61\% and an average rank of \textbf{2.59}, outperforming the recent state-of-the-art SoftShape~\cite{softshape} by 1.35\% in accuracy. Notably, PDFTime achieves the top-1 accuracy on \textbf{80} out of 128 datasets, a margin that confirms its exceptional generalization across highly diverse temporal patterns.
\begin{table}[t]
\centering
\caption{Ablation study on embedding and prototype design.
Results are averaged over 10 UEA datasets.}
\label{tab:ablation_embedding}
\begin{tabular}{lc}
\toprule
\textbf{Variant} 
& \textbf{Avg. Acc (\%)}  \\
\midrule
w/o FFT weight  & 0.766  \\
Linear embedding  & 0.686 \\
\midrule
\textbf{PDFTime (Full)}          & \textbf{0.783} \\
\bottomrule
\end{tabular}
\vspace{-5mm}
\end{table}
\begin{table}[t]
\centering
\caption{Ablation study on embedding and prototype design.
Results are averaged over 10 UEA datasets.}
\label{tab:ablation_prototype}
\begin{tabular}{lc}
\toprule
\textbf{Variant} 
& \textbf{Avg. Acc (\%)} \\
\midrule
Linear Head  & 0.754  \\
One field Prototype Head                & 0.763 \\
back progation       & 0.761 \\

\midrule
\textbf{PDFTime (Full)}          & \textbf{0.783} \\
\bottomrule
\end{tabular}
\vspace{-3mm}
\end{table}
\begin{table}[htbp]
\centering
\caption{Effect of prototype update strategy ($\gamma$).
Results are averaged over 10 UEA datasets.}
\label{tab:ablation_gamma}
\begin{tabular}{lc}
\toprule
\textbf{Update Strategy} 
& \textbf{Avg. Acc (\%)}  \\
\midrule
stable $\gamma = 1$    & 0.767 \\
stable $\gamma = 0.999$    & 0.771 \\
stable $\gamma = 0.95$            & 0.768 \\
\midrule
\textbf{PDFTime(Adaptive $\gamma$)} & \textbf{0.783} \\
\bottomrule
\end{tabular}
\end{table}

\vspace{-2mm}

\subsection{Ablation Experiments}

\begin{figure*}[t]
    \centering
    \includegraphics[width=0.7\linewidth]{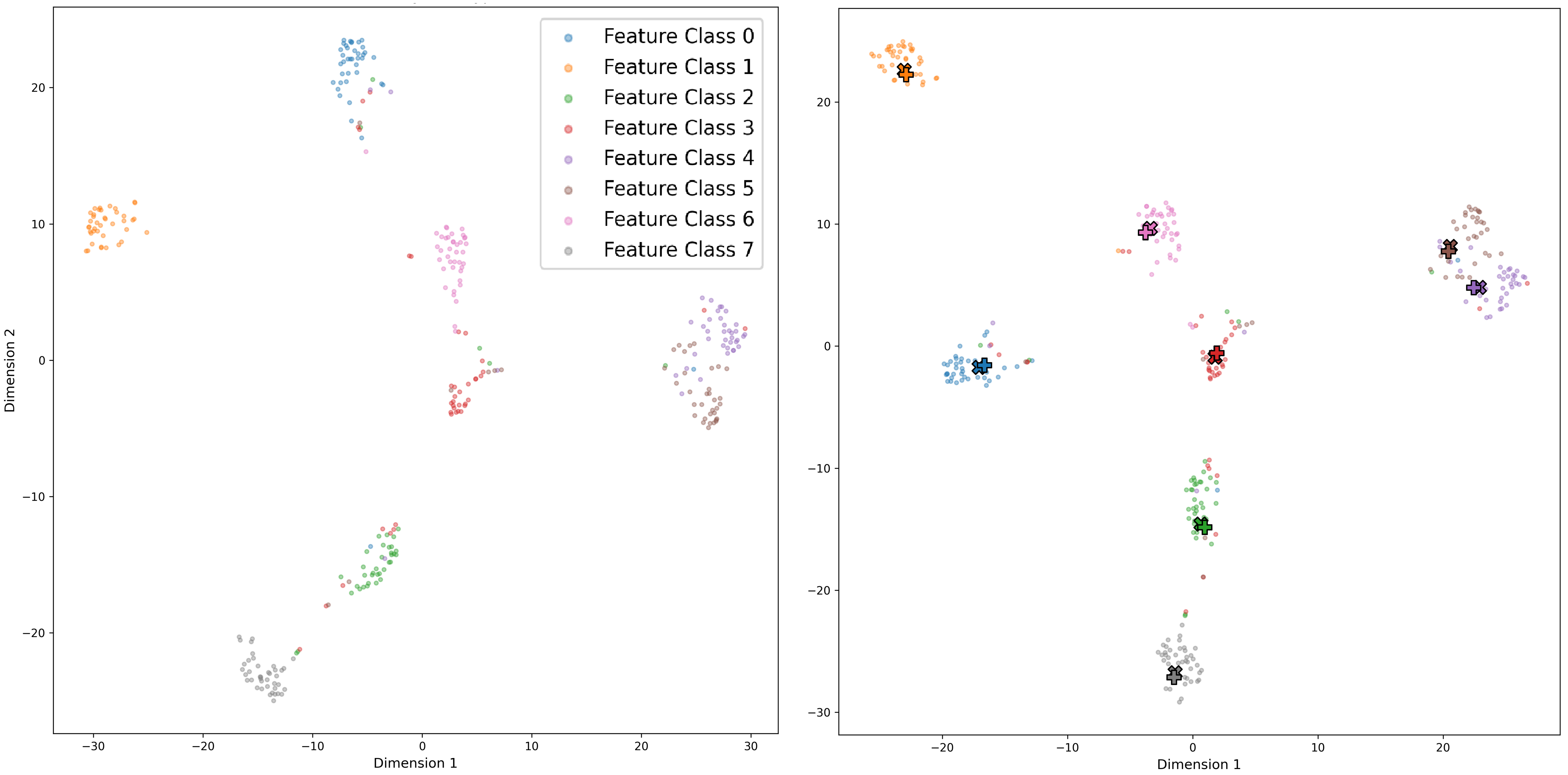}
    \caption{T-SNE on UWaveGestureLibrary. The left figure is from the MLP classification head, and the right figure is ours. The prototype is represented by X in the diagram.}
    \label{fig:t-sne}  
\end{figure*}

\begin{figure*}[t]
    \centering
    \includegraphics[width=0.7\linewidth]{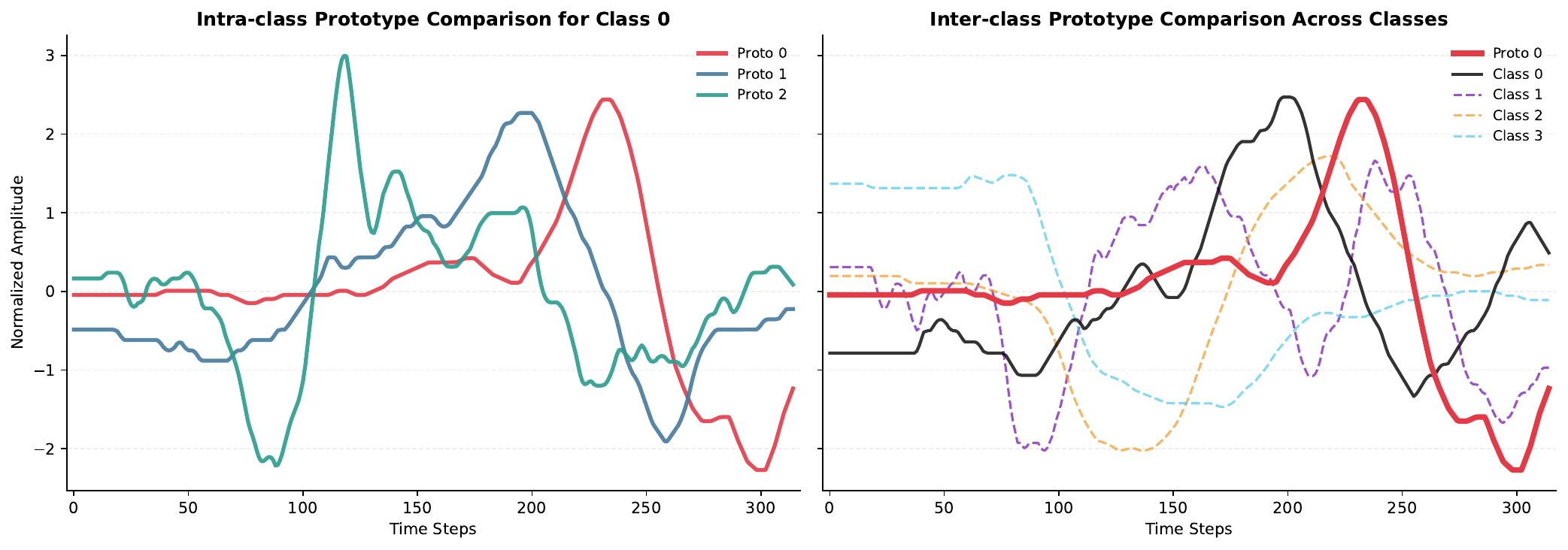}
    \caption{Comparison between learned prototypes and original time series on the UWaveGestureLibrary dataset.}
    \label{fig:moti}  
    \vspace{-3mm}
\end{figure*}

\textbf{Effect of Prototype-Based Reasoning.}

We evaluate the contribution of our hierarchical design in Table~\ref{tab:ablation_prototype}. The full PDFTime consistently outperforms both prototype-free and single-level variants. This validates the benefit of progressive similarity aggregation—moving from fine-grained temporal patterns to consolidated class-level decisions.

\textbf{Effect of Prototype-Based Classification Head.}
To isolate the impact of the decision mechanism, we compare our prototype head against a standard linear head using identical feature representations. As shown in Table~\ref{tab:ablation_gamma}, the prototype head yields superior performance, proving that the gains are not merely due to backbone capacity but stem from the explicit similarity-based reasoning which provides a more robust inductive bias than flat linear projections.

\textbf{Effect of the Prototype Update Strategy ($\gamma$).}

We analyze the influence of momentum scheduling in Table~\ref{tab:ablation_gamma}. While fixed prototypes ($\gamma=1$) offer a decent baseline due to our structured QR-based initialization, allowing adaptation is crucial. However, rapid updates (e.g., $\gamma_a=0.7$) destabilize the latent space. Our adaptive scheduling (Alg. 1) strikes the optimal balance between initial stability and late-stage convergence.

\subsection{Prototype Interpretability} 
To examine the model's underlying decision logic, we project learned embeddings and prototypes into a 2D space via t-SNE (Fig.~\ref{fig:t-sne}), complemented by temporal motif alignment (Fig.~\ref{fig:moti}) which maps abstract prototypes back to physical signals. The visualizations yield two key insights:

First, prototypes act as representative centroids that align precisely with the latent data distribution. By consistently occupying high-density regions within clusters, these prototypes serve as transparent evidence for the model's reasoning. Consequently, classification is reformulated from a black-box mapping into an explicit attribution process, where a prediction is justified by its similarity to the most relevant prototypical motifs.

Second, the framework demonstrates structured diversity. Multiple prototypes within a single class capture distinct intra-class temporal variants, while maintaining a robust discriminative margin against other classes. This hierarchical organization allows PDFTime to resolve complex global tasks through a series of simpler, similarity-driven sub-tasks.

Together, these properties provide PDFTime with intrinsic transparency. By grounding high-dimensional features in human-understandable temporal patterns, the model offers a direct and lossless link between latent reasoning and physical signals, eliminating the need for unreliable post-hoc explanation tools. Further case studies are in Appendix~\ref{app:visual}.

\vspace{-2mm}
\section{Conclusion}
This paper addresses the limited interpretability of existing time series classification methods by proposing PDFTime, a multi-granularity prototype-guided framework that decouples representation learning from decision making to enable transparent, prototype-based reasoning while preserving deep model expressiveness.
Extensive subject-independent experiments show that PDFTime consistently outperforms ten representative baselines across multiple datasets, demonstrating strong performance, robustness, and generalization in challenging real-world scenarios.

\section*{Impact Statement}

This paper presents a prototype-guided framework for time series classification with the goal of improving generalization and interpretability in deep learning models. Time series classification is a foundational task in machine learning with established applications in areas such as healthcare monitoring, industrial systems, and scientific data analysis.

The proposed method focuses on model architecture and learning mechanisms and does not introduce new data sources, user profiling techniques, or decision-making policies that could pose direct ethical concerns. By enhancing interpretability through structured, similarity-based reasoning, the framework may support more transparent and trustworthy deployment of time series models in high-stakes scenarios.

As with most advances in machine learning methodology, potential societal impacts depend on the specific application context and deployment practices. We do not foresee any negative ethical implications that require special consideration beyond those commonly associated with machine learning research. Overall, this work aims to advance the methodological foundations of time series classification.





\bibliography{ref}
\bibliographystyle{icml2025}

\newpage
\appendix
\onecolumn
\section{Experimental Setup}

\subsection{UEA Time Series Classification Benchmark}
To assess performance on multivariate tasks, we incorporate datasets from the University of East Anglia (UEA) repository. We evaluate our framework on 10 preprocessed datasets covering real-world scenarios, including human activity recognition and sensor data interpretation. These datasets offer a multidimensional perspective, requiring the model to capture complex inter-variable correlations. Detailed statistics for Datasets in UEA benchmark are provided in Table~\ref{tab:uea_data}.
\begin{table}[htbp]
\centering
\caption{Summary of the 10 datasets in UEA benchmark.}
\label{tab:uea_data}
\begin{tabular}{l|rrrrr}
\toprule
Dataset & Train Size & Test Size & Dimensions & Length & Classes \\
\midrule
EthanolConcentration      & 261 & 263 & 3 & 1751 & 4  \\
FaceDetection             & 5890 & 3524 & 144 & 62  & 2  \\
Handwriting               & 150 & 850 & 3 & 152 & 26 \\
Heartbeat                 & 204 & 205 & 61 & 405 & 2  \\
JapaneseVowels            & 270 & 370 & 12 & 29 (max) & 9  \\
PEMS-SF                   & 267 & 173 & 963 & 144 & 7  \\
SelfRegulationSCP1        & 268 & 293 & 6 & 896 & 2  \\
SelfRegulationSCP2        & 200 & 180 & 7 & 1152 & 2  \\
SpokenArabicDigits        & 6599 & 2199 & 13 & 93 (max) & 10 \\
UWaveGestureLibrary       & 120 & 320 & 3 & 315 & 8  \\
\bottomrule
\end{tabular}
\end{table}

\subsection{UCR Time Series Classification Benchmark}

The UCR Time Series Classification Archive \cite{ucr} is one of the most comprehensive collections of univariate datasets for time series analysis. It encompasses 128 datasets spanning diverse domains such as healthcare, finance, and environmental monitoring. The variety within this archive provides a robust platform to evaluate the effectiveness and generalization of PDFTime. Notably, several datasets in the archive (e.g., CBF and DiatomSizeReduction) exhibit a substantial imbalance between training and test sample sizes, posing additional challenges for model generalization. Since the original UCR archive lacks predefined validation sets, we strictly follow the experimental protocols in \cite{timesnet, wang2024tssurvey} to ensure fair comparisons and avoid data leakage. Detailed statistics for Datasets in UCR benchmark are provided in Table~\ref{tab:ucr_128_datasets}. All results in Table~\ref{tab:ucr_all}.
\begin{table}[!htbp]
\centering
\caption{Summary of the 128 UCR Time Series Classification Datasets}
\label{tab:ucr_128_datasets}
\resizebox{\linewidth}{!}{
\begin{tabular}{r l r r r r r | r l r r r r r}
\toprule
ID & Name & \#Train & \#Test & \#Total & \#Class & Length &
ID & Name & \#Train & \#Test & \#Total & \#Class & Length \\
\midrule
1 & ACSF1 & 100 & 100 & 200 & 10 & 1460 &
65 & ItalyPowerDemand & 67 & 1029 & 1096 & 2 & 24 \\
2 & Adiac & 390 & 391 & 781 & 37 & 176 &
66 & LargeKitchenAppliances & 375 & 375 & 750 & 3 & 720 \\
3 & AllGestureWiimoteX & 300 & 700 & 1000 & 10 & Vary &
67 & Lightning2 & 60 & 61 & 121 & 2 & 637 \\
4 & AllGestureWiimoteY & 300 & 700 & 1000 & 10 & Vary &
68 & Lightning7 & 70 & 73 & 143 & 7 & 319 \\
5 & AllGestureWiimoteZ & 300 & 700 & 1000 & 10 & Vary &
69 & Mallat & 55 & 2345 & 2400 & 8 & 1024 \\
6 & ArrowHead & 36 & 175 & 211 & 3 & 251 &
70 & Meat & 60 & 60 & 120 & 3 & 448 \\
7 & Beef & 30 & 30 & 60 & 5 & 470 &
71 & MedicalImages & 381 & 760 & 1141 & 10 & 99 \\
8 & BeetleFly & 20 & 20 & 40 & 2 & 512 &
72 & MelbournePedestrian & 1194 & 2439 & 3633 & 10 & 24 \\
9 & BirdChicken & 20 & 20 & 40 & 2 & 512 &
73 & MiddlePhalanxOutlineAgeGroup & 400 & 154 & 554 & 3 & 80 \\
10 & BME & 30 & 150 & 180 & 3 & 128 &
74 & MiddlePhalanxOutlineCorrect & 600 & 291 & 891 & 2 & 80 \\
11 & Car & 60 & 60 & 120 & 4 & 577 &
75 & MiddlePhalanxTW & 399 & 154 & 553 & 6 & 80 \\
12 & CBF & 30 & 900 & 930 & 3 & 128 &
76 & MixedShapesRegularTrain & 500 & 2425 & 2925 & 5 & 1024 \\
13 & Chinatown & 20 & 343 & 363 & 2 & 24 &
77 & MixedShapesSmallTrain & 100 & 2425 & 2525 & 5 & 1024 \\
14 & ChlorineConcentration & 467 & 3840 & 4307 & 3 & 166 &
78 & MoteStrain & 20 & 1252 & 1272 & 2 & 84 \\
15 & CinCECGTorso & 40 & 1380 & 1420 & 4 & 1639 &
79 & NonInvasiveFetalECGThorax1 & 1800 & 1965 & 3765 & 42 & 750 \\
16 & Coffee & 28 & 28 & 56 & 2 & 286 &
80 & NonInvasiveFetalECGThorax2 & 1800 & 1965 & 3765 & 42 & 750 \\
17 & Computers & 250 & 250 & 500 & 2 & 720 &
81 & OliveOil & 30 & 30 & 60 & 4 & 570 \\
18 & CricketX & 390 & 390 & 780 & 12 & 300 &
82 & OSULeaf & 200 & 242 & 442 & 6 & 427 \\
19 & CricketY & 390 & 390 & 780 & 12 & 300 &
83 & PhalangesOutlinesCorrect & 1800 & 858 & 2658 & 2 & 80 \\
20 & CricketZ & 390 & 390 & 780 & 12 & 300 &
84 & Phoneme & 214 & 1896 & 2110 & 39 & 1024 \\
21 & Crop & 7200 & 16800 & 24000 & 24 & 46 &
85 & PickupGestureWiimoteZ & 50 & 50 & 100 & 10 & Vary \\
22 & DiatomSizeReduction & 16 & 306 & 322 & 4 & 345 &
86 & PigAirwayPressure & 104 & 208 & 312 & 52 & 2000 \\
23 & DistalPhalanxOutlineAgeGroup & 400 & 139 & 539 & 3 & 80 &
87 & PigArtPressure & 104 & 208 & 312 & 52 & 2000 \\
24 & DistalPhalanxOutlineCorrect & 600 & 276 & 876 & 2 & 80 &
88 & PigCVP & 104 & 208 & 312 & 52 & 2000 \\
25 & DistalPhalanxTW & 400 & 139 & 539 & 6 & 80 &
89 & PLAID & 537 & 537 & 1074 & 11 & Vary \\
26 & DodgerLoopDay & 78 & 80 & 158 & 7 & 288 &
90 & Plane & 105 & 105 & 210 & 7 & 144 \\
27 & DodgerLoopGame & 20 & 138 & 158 & 2 & 288 &
91 & PowerCons & 180 & 180 & 360 & 2 & 144 \\
28 & DodgerLoopWeekend & 20 & 138 & 158 & 2 & 288 &
92 & ProximalPhalanxOutlineAgeGroup & 400 & 205 & 605 & 3 & 80 \\
29 & Earthquakes & 322 & 139 & 461 & 2 & 512 &
93 & ProximalPhalanxOutlineCorrect & 600 & 291 & 891 & 2 & 80 \\
30 & ECG200 & 100 & 100 & 200 & 2 & 96 &
94 & ProximalPhalanxTW & 400 & 205 & 605 & 6 & 80 \\
31 & ECG5000 & 500 & 4500 & 5000 & 5 & 140 &
95 & RefrigerationDevices & 375 & 375 & 750 & 3 & 720 \\
32 & ECGFiveDays & 23 & 861 & 884 & 2 & 136 &
96 & Rock & 20 & 50 & 70 & 4 & 2844 \\
33 & ElectricDevices & 8926 & 7711 & 16637 & 7 & 96 &
97 & ScreenType & 375 & 375 & 750 & 3 & 720 \\
34 & EOGHorizontalSignal & 362 & 362 & 724 & 12 & 1250 &
98 & SemgHandGenderCh2 & 300 & 600 & 900 & 2 & 1500 \\
35 & EOGVerticalSignal & 362 & 362 & 724 & 12 & 1250 &
99 & SemgHandMovementCh2 & 450 & 450 & 900 & 6 & 1500 \\
36 & EthanolLevel & 504 & 500 & 1004 & 4 & 1751 &
100 & SemgHandSubjectCh2 & 450 & 450 & 900 & 5 & 1500 \\
37 & FaceAll & 560 & 1690 & 2250 & 14 & 131 &
101 & ShakeGestureWiimoteZ & 50 & 50 & 100 & 10 & Vary \\
38 & FaceFour & 24 & 88 & 112 & 4 & 350 &
102 & ShapeletSim & 20 & 180 & 200 & 2 & 500 \\
39 & FaceUCR & 200 & 2050 & 2250 & 14 & 131 &
103 & ShapesAll & 600 & 600 & 1200 & 60 & 512 \\
40 & FiftyWords & 450 & 455 & 905 & 50 & 270 &
104 & SmallKitchenAppliances & 375 & 375 & 750 & 3 & 720 \\
41 & Fish & 175 & 175 & 350 & 7 & 463 &
105 & SmoothSubspace & 150 & 150 & 300 & 3 & 15 \\
42 & FordA & 3601 & 1320 & 4921 & 2 & 500 &
106 & SonyAIBORobotSurface1 & 20 & 601 & 621 & 2 & 70 \\
43 & FordB & 3636 & 810 & 4446 & 2 & 500 &
107 & SonyAIBORobotSurface2 & 27 & 953 & 980 & 2 & 65 \\
44 & FreezerRegularTrain & 150 & 2850 & 3000 & 2 & 301 &
108 & StarLightCurves & 1000 & 8236 & 9236 & 3 & 1024 \\
45 & FreezerSmallTrain & 28 & 2850 & 2878 & 2 & 301 &
109 & Strawberry & 613 & 370 & 983 & 2 & 235 \\
46 & Fungi & 18 & 186 & 204 & 18 & 201 &
110 & SwedishLeaf & 500 & 625 & 1125 & 15 & 128 \\
47 & GestureMidAirD1 & 208 & 130 & 338 & 26 & Vary &
111 & Symbols & 25 & 995 & 1020 & 6 & 398 \\
48 & GestureMidAirD2 & 208 & 130 & 338 & 26 & Vary &
112 & SyntheticControl & 300 & 300 & 600 & 6 & 60 \\
49 & GestureMidAirD3 & 208 & 130 & 338 & 26 & Vary &
113 & ToeSegmentation1 & 40 & 228 & 268 & 2 & 277 \\
50 & GesturePebbleZ1 & 132 & 172 & 304 & 6 & Vary &
114 & ToeSegmentation2 & 36 & 130 & 166 & 2 & 343 \\
51 & GesturePebbleZ2 & 146 & 158 & 304 & 6 & Vary &
115 & Trace & 100 & 100 & 200 & 4 & 275 \\
52 & GunPoint & 50 & 150 & 200 & 2 & 150 &
116 & TwoLeadECG & 23 & 1139 & 1162 & 2 & 82 \\
53 & GunPointAgeSpan & 135 & 316 & 451 & 2 & 150 &
117 & TwoPatterns & 1000 & 4000 & 5000 & 4 & 128 \\
54 & GunPointMaleVersusFemale & 135 & 316 & 451 & 2 & 150 &
118 & UMD & 36 & 144 & 180 & 3 & 150 \\
55 & GunPointOldVersusYoung & 136 & 315 & 451 & 2 & 150 &
119 & UWaveGestureLibraryAll & 896 & 3582 & 4478 & 8 & 945 \\
56 & Ham & 109 & 105 & 214 & 2 & 431 &
120 & UWaveGestureLibraryX & 896 & 3582 & 4478 & 8 & 315 \\
57 & HandOutlines & 1000 & 370 & 1370 & 2 & 2709 &
121 & UWaveGestureLibraryY & 896 & 3582 & 4478 & 8 & 315 \\
58 & Haptics & 155 & 308 & 463 & 5 & 1092 &
122 & UWaveGestureLibraryZ & 896 & 3582 & 4478 & 8 & 315 \\
59 & Herring & 64 & 64 & 128 & 2 & 512 &
123 & Wafer & 1000 & 6164 & 7164 & 2 & 152 \\
60 & HouseTwenty & 40 & 119 & 159 & 2 & 2000 &
124 & Wine & 57 & 54 & 111 & 2 & 234 \\
61 & InlineSkate & 100 & 550 & 650 & 7 & 1882 &
125 & WordSynonyms & 267 & 638 & 905 & 25 & 270 \\
62 & InsectEPGRegularTrain & 62 & 249 & 311 & 3 & 601 &
126 & Worms & 181 & 77 & 258 & 5 & 900 \\
63 & InsectEPGSmallTrain & 17 & 249 & 266 & 3 & 601 &
127 & WormsTwoClass & 181 & 77 & 258 & 2 & 900 \\
64 & InsectWingbeatSound & 220 & 1980 & 2200 & 11 & 256 &
128 & Yoga & 300 & 3000 & 3300 & 2 & 426 \\
\bottomrule
\end{tabular}}
\end{table}

\begin{table*}[!htbp]
\centering
\caption{All results on UCR time series datasets. Best results are highlighted in red, and second-best results in blue.}
\label{tab:ucr_all}
\resizebox{1\textwidth}{!}{
\begin{tabular}{lccccccccc}
\toprule
\textbf{Datasets} &
\textbf{TEST} &
\textbf{patchTST} &
\textbf{TSLANET} &
\textbf{SARC} &
\textbf{softshape} &
\textbf{TimesNet} &
\textbf{ROCKET} &
\textbf{InceptionTime} &
\textbf{PDFTime(Ours)} \\
\midrule
ACSF1 & 0.7760 & 0.7600 & 0.8700 & 0.8500 & \textcolor{blue}{0.9100} & 0.6000 & 0.8700 & 0.8300 & \textcolor{red}{0.9900} \\
Adiac & 0.8250 & 0.8358 & 0.8837 & 0.7683 & \textcolor{blue}{0.9374} & 0.8082 & 0.7754 & 0.7673 & \textcolor{red}{0.9603} \\
AllGestureWiimoteX & 0.7660 & 0.5730 & 0.8970 & 0.7234 & \textcolor{blue}{0.8980} & 0.5120 & 0.7771 & 0.7600 & \textcolor{red}{1.0000} \\
AllGestureWiimoteY & 0.8530 & 0.6730 & \textcolor{red}{0.9180} & 0.7583 & \textcolor{blue}{0.9110} & 0.5820 & 0.7549 & 0.7783 & 0.9050 \\
AllGestureWiimoteZ & 0.8080 & 0.5890 & \textcolor{blue}{0.9010} & 0.7474 & 0.8850 & 0.6620 & 0.7494 & 0.7797 & \textcolor{red}{0.9890} \\
ArrowHead & 0.8830 & 0.7078 & \textcolor{blue}{0.9672} & 0.8537 & 0.9435 & 0.7821 & 0.8011 & 0.8480 & \textcolor{red}{1.0000} \\
Beef & \textcolor{red}{1.0000} & 0.4167 & 0.7500 & \textcolor{blue}{0.9133} & 0.8667 & 0.8000 & 0.8000 & 0.6600 & \textcolor{red}{1.0000} \\
BeetleFly & 0.8100 & 0.8500 & \textcolor{blue}{0.9750} & 0.9500 & \textcolor{blue}{0.9750} & 0.7000 & 0.9000 & 0.8500 & \textcolor{red}{1.0000} \\
BirdChicken & 0.8150 & 0.8500 & 0.8750 & \textcolor{blue}{0.9400} & 0.9000 & 0.8250 & 0.9000 & 0.8000 & \textcolor{red}{1.0000} \\
BME & \textcolor{red}{1.0000} & 0.9500 & \textcolor{red}{1.0000} & \textcolor{red}{1.0000} & \textcolor{blue}{0.9944} & 0.9778 & \textcolor{red}{1.0000} & 0.9933 & \textcolor{red}{1.0000} \\
Car & 0.6320 & 0.9083 & 0.9500 & 0.9267 & \textcolor{blue}{0.9583} & 0.8417 & 0.8833 & 0.9000 & \textcolor{red}{1.0000} \\
CBF & 0.8020 & \textcolor{red}{1.0000} & \textcolor{red}{1.0000} & 0.9882 & \textcolor{red}{1.0000} & \textcolor{red}{1.0000} & \textcolor{blue}{0.9989} & 0.9956 & 0.9972 \\
Chinatown & 0.7540 & 0.9836 & 0.9863 & 0.9808 & \textcolor{blue}{0.9890} & 0.9863 & 0.9796 & 0.9825 & \textcolor{red}{1.0000} \\
ChlorineConcentration & 0.7870 & \textcolor{red}{0.9995} & 0.9988 & 0.8672 & 0.9988 & \textcolor{blue}{0.9993} & 0.8061 & 0.8616 & 0.5668 \\
CinCECGTorso & 0.9800 & \textcolor{red}{1.0000} & \textcolor{red}{1.0000} & 0.9468 & \textcolor{red}{1.0000} & \textcolor{blue}{0.9930} & 0.8275 & 0.8239 & \textcolor{red}{1.0000} \\
Coffee & \textcolor{blue}{0.7760} & \textcolor{red}{1.0000} & \textcolor{red}{1.0000} & \textcolor{red}{1.0000} & \textcolor{red}{1.0000} & \textcolor{red}{1.0000} & \textcolor{red}{1.0000} & \textcolor{red}{1.0000} & \textcolor{red}{1.0000} \\
Computers & 0.7140 & 0.8380 & \textcolor{blue}{0.8920} & 0.7936 & 0.8160 & 0.8120 & 0.7720 & 0.7880 & \textcolor{red}{0.9600} \\
CricketX & 0.6620 & 0.7821 & 0.9205 & 0.8036 & \textcolor{blue}{0.9321} & 0.7974 & 0.8164 & 0.7451 & \textcolor{red}{0.9885} \\
CricketY & 0.7460 & 0.7821 & 0.8987 & 0.7887 & \textcolor{blue}{0.9154} & 0.7808 & 0.8421 & 0.8231 & \textcolor{red}{0.9308} \\
CricketZ & 0.8930 & 0.7923 & \textcolor{blue}{0.9064} & 0.8097 & \textcolor{red}{0.9372} & 0.7962 & 0.8467 & 0.8354 & 0.8936 \\
Crop & \textcolor{blue}{0.9350} & 0.1193 & 0.5781 & 0.7238 & 0.8765 & 0.8503 & 0.7543 & 0.7608 & \textcolor{red}{0.9872} \\
DiatomSizeReduction & \textcolor{red}{1.0000} & \textcolor{red}{1.0000} & \textcolor{blue}{0.9969} & 0.9843 & \textcolor{red}{1.0000} & 0.9938 & 0.9673 & 0.9837 & \textcolor{red}{1.0000} \\
DistalPhalanxOutlineAgeGroup & 0.7140 & 0.8779 & \textcolor{blue}{0.9333} & 0.7612 & \textcolor{red}{0.9518} & 0.8590 & 0.7410 & 0.7511 & 0.9091 \\
DistalPhalanxOutlineCorrect & 0.7890 & 0.9009 & 0.9018 & 0.7746 & \textcolor{blue}{0.9054} & 0.8860 & 0.7645 & 0.7746 & \textcolor{red}{0.9441} \\
DistalPhalanxTW & 0.8340 & 0.8485 & \textcolor{blue}{0.8796} & 0.7396 & \textcolor{red}{0.9037} & 0.8499 & 0.6777 & 0.6978 & 0.7959 \\
DodgerLoopDay & \textcolor{blue}{0.9390} & 0.8485 & 0.8244 & 0.5875 & 0.8623 & 0.7107 & 0.5925 & 0.5175 & \textcolor{red}{1.0000} \\
DodgerLoopGame & 0.7810 & 0.6694 & \textcolor{blue}{0.9563} & 0.9304 & 0.9433 & 0.8484 & 0.8406 & 0.8551 & \textcolor{red}{0.9937} \\
DodgerLoopWeekend & 0.9370 & 0.8872 & \textcolor{blue}{0.9873} & 0.9826 & \textcolor{red}{1.0000} & 0.9688 & 0.9710 & 0.9565 & \textcolor{red}{1.0000} \\
Earthquakes & 0.9400 & \textcolor{blue}{0.9492} & 0.9222 & 0.7568 & 0.8917 & 0.9136 & 0.7424 & 0.7410 & \textcolor{red}{0.9800} \\
ECG200 & 0.7890 & 0.9115 & 0.9450 & 0.8980 & \textcolor{blue}{0.9475} & 0.9100 & 0.9100 & 0.9000 & \textcolor{red}{0.9502} \\
ECG5000 & \textcolor{blue}{0.9830} & 0.8763 & 0.9760 & 0.9456 & 0.9796 & 0.9708 & 0.9460 & 0.9398 & \textcolor{red}{1.0000} \\
ECGFiveDays & 0.7140 & \textcolor{blue}{0.9364} & \textcolor{red}{1.0000} & \textcolor{red}{1.0000} & \textcolor{red}{1.0000} & \textcolor{red}{1.0000} & \textcolor{red}{1.0000} & \textcolor{red}{1.0000} & 0.9185 \\
ElectricDevices & 0.9180 & \textcolor{red}{1.0000} & 0.8991 & 0.7387 & \textcolor{blue}{0.9208} & 0.8610 & 0.7268 & 0.7189 & 0.7983 \\
EOGHorizontalSignal & 0.5100 & 0.4883 & 0.8993 & 0.6155 & \textcolor{blue}{0.9007} & 0.6092 & 0.6177 & 0.5663 & \textcolor{red}{0.9426} \\
EOGVerticalSignal & 0.6250 & 0.8010 & \textcolor{red}{0.8965} & 0.5271 & \textcolor{blue}{0.8814} & 0.6369 & 0.5365 & 0.4762 & 0.8439 \\
EthanolLevel & 0.3890 & 0.8013 & 0.9184 & 0.6036 & \textcolor{blue}{0.9204} & 0.7690 & 0.5868 & 0.7700 & \textcolor{red}{1.0000} \\
FaceAll & 0.6200 & 0.7248 & 0.9947 & 0.9140 & \textcolor{blue}{0.9978} & 0.9733 & 0.9462 & 0.8130 & \textcolor{red}{1.0000} \\
FaceFour & 0.9690 & 0.8057 & \textcolor{blue}{0.9913} & 0.9341 & 0.9909 & 0.9557 & 0.9773 & 0.9659 & \textcolor{red}{1.0000} \\
FacesUCR & 0.8550 & 0.9068 & \textcolor{blue}{0.9960} & 0.9410 & 0.9956 & 0.9800 & 0.9611 & 0.9634 & \textcolor{red}{1.0000} \\
FiftyWords & 0.8460 & \textcolor{blue}{0.9824} & 0.9249 & 0.7776 & 0.9359 & 0.8276 & 0.8290 & 0.8044 & \textcolor{red}{1.0000} \\
Fish & 0.8660 & 0.8417 & 0.9629 & 0.9646 & 0.9771 & 0.8000 & 0.9829 & \textcolor{blue}{0.9851} & \textcolor{red}{0.9902} \\
FordA & 0.9150 & 0.9343 & 0.9740 & 0.9561 & \textcolor{blue}{0.9772} & 0.9348 & 0.9353 & 0.9524 & \textcolor{red}{0.9780} \\
FordB & 0.9500 & 0.9500 & 0.9658 & 0.8398 & \textcolor{blue}{0.9744} & 0.9046 & 0.7993 & 0.8432 & \textcolor{red}{0.9997} \\
FreezerRegularTrain & 0.7920 & 0.9377 & \textcolor{red}{0.9993} & 0.9959 & \textcolor{blue}{0.9990} & \textcolor{blue}{0.9990} & 0.9973 & 0.9959 & \textcolor{red}{0.9993} \\
FreezerSmallTrain & 0.8110 & 0.9393 & \textcolor{blue}{0.9997} & 0.9254 & 0.9990 & 0.9969 & 0.9478 & 0.8406 & \textcolor{red}{1.0000} \\
Fungi & 0.6360 & \textcolor{blue}{0.8492} & \textcolor{red}{1.0000} & \textcolor{red}{1.0000} & \textcolor{red}{1.0000} & 0.8337 & \textcolor{red}{1.0000} & \textcolor{red}{1.0000} & \textcolor{red}{1.0000} \\
GestureMidAirD1 & 0.5910 & 0.7276 & 0.8493 & 0.7600 & \textcolor{blue}{0.8640} & 0.6099 & 0.7231 & 0.7385 & \textcolor{red}{0.9438} \\
GestureMidAirD2 & \textcolor{blue}{0.8570} & 0.6103 & 0.7640 & 0.6877 & 0.7962 & 0.5449 & 0.6692 & 0.6846 & \textcolor{red}{0.9911} \\
GestureMidAirD3 & \textcolor{blue}{0.9230} & 0.5692 & 0.6577 & 0.4615 & 0.7613 & 0.4559 & 0.3938 & 0.3923 & \textcolor{red}{1.0000} \\
GesturePebbleZ1 & 0.9400 & 0.4962 & \textcolor{blue}{0.9967} & 0.8837 & 0.9867 & 0.9541 & 0.9070 & 0.9291 & \textcolor{red}{1.0000} \\
GesturePebbleZ2 & 0.9030 & 0.8316 & 0.9738 & 0.8266 & \textcolor{blue}{0.9836} & 0.8850 & 0.8241 & 0.8633 & \textcolor{red}{1.0000} \\
GunPoint & 0.8720 & 0.7800 & 0.9900 & \textcolor{red}{1.0000} & 0.9900 & 0.9850 & \textcolor{red}{1.0000} & 0.9933 & \textcolor{blue}{0.9956} \\
GunPointAgeSpan & 0.7940 & 0.9800 & 0.9890 & 0.9956 & 0.9868 & 0.9823 & \textcolor{blue}{0.9968} & 0.9937 & \textcolor{red}{1.0000} \\
GunPointMaleVersusFemale & 0.2960 & 0.9889 & 0.9934 & \textcolor{red}{1.0000} & 0.9956 & 0.9956 & 0.9968 & 0.9968 & \textcolor{blue}{0.9978} \\
GunPointOldVersusYoung & \textcolor{red}{1.0000} & \textcolor{red}{1.0000} & 0.9823 & \textcolor{blue}{0.9981} & 0.9779 & 0.9800 & 0.9879 & 0.9937 & 0.9953 \\
Ham & 0.8760 & \textcolor{red}{0.9912} & 0.8972 & 0.7505 & \textcolor{blue}{0.9161} & 0.9023 & 0.7143 & 0.7810 & 0.7533 \\
HandOutlines & 0.8440 & 0.5420 & 0.9460 & 0.9470 & \textcolor{red}{0.9533} & 0.9029 & 0.9405 & \textcolor{blue}{0.9508} & 0.9375 \\
Haptics & 0.7850 & \textcolor{red}{0.9350} & 0.6961 & 0.5773 & 0.7522 & 0.6422 & 0.5266 & 0.5519 & \textcolor{blue}{0.9330} \\
Herring & 0.5870 & 0.6672 & \textcolor{blue}{0.8212} & 0.6188 & 0.7671 & 0.6560 & 0.5938 & 0.7031 & \textcolor{red}{1.0000} \\
HouseTwenty & 0.4050 & 0.7604 & 0.9500 & 0.9630 & \textcolor{blue}{0.9688} & 0.8615 & 0.9580 & 0.9496 & \textcolor{red}{1.0000} \\
InlineSkate & \textcolor{red}{0.9890} & \textcolor{blue}{0.8617} & 0.7769 & 0.5044 & 0.8077 & 0.5369 & 0.4662 & 0.4818 & 0.6754 \\
InsectEPGRegularTrain & 0.9330 & 0.8985 & 0.9904 & \textcolor{red}{1.0000} & \textcolor{blue}{0.9968} & 0.8267 & \textcolor{red}{1.0000} & \textcolor{red}{1.0000} & \textcolor{red}{1.0000} \\
InsectEPGSmallTrain & 0.9150 & 0.9617 & 0.9813 & \textcolor{blue}{0.9831} & \textcolor{red}{0.9963} & 0.8011 & 0.9799 & 0.8594 & 0.8300 \\
InsectWingbeatSound & \textcolor{red}{1.0000} & 0.9511 & 0.8786 & 0.6542 & 0.8882 & 0.7355 & 0.6599 & 0.6157 & \textcolor{blue}{0.9754} \\
ItalyPowerDemand & 0.9820 & 0.8609 & \textcolor{red}{0.9891} & 0.9668 & \textcolor{red}{0.9891} & \textcolor{blue}{0.9863} & 0.9689 & 0.9673 & 0.9813 \\
LargeKitchenAppliances & \textcolor{red}{1.0000} & \textcolor{blue}{0.9827} & 0.9533 & 0.8901 & 0.9627 & 0.7520 & 0.8965 & 0.8805 & \textcolor{red}{1.0000} \\
Lightning2 & 0.8100 & 0.8253 & \textcolor{blue}{0.9100} & 0.8197 & 0.8857 & 0.8937 & 0.7738 & 0.8197 & \textcolor{red}{1.0000} \\
Lightning7 & 0.7290 & 0.8219 & 0.8892 & 0.8274 & \textcolor{blue}{0.8966} & 0.8264 & 0.8082 & 0.7452 & \textcolor{red}{0.9988} \\
Mallat & 0.7610 & 0.7025 & \textcolor{red}{1.0000} & 0.9597 & \textcolor{blue}{0.9988} & 0.9954 & 0.9557 & 0.8936 & \textcolor{red}{1.0000} \\
Meat & 0.9350 & \textcolor{blue}{0.9996} & \textcolor{red}{1.0000} & 0.9800 & \textcolor{red}{1.0000} & \textcolor{red}{1.0000} & 0.9500 & 0.9300 & 0.8273 \\
\bottomrule
\end{tabular}
}
\end{table*}

\begin{table*}[htbp]
\centering
\resizebox{1\textwidth}{!}{
\begin{tabular}{lccccccccc}
\toprule
\textbf{Datasets} &
\textbf{TEST} &
\textbf{patchTST} &
\textbf{TSLANET} &
\textbf{SARC} &
\textbf{softshape} &
\textbf{TimesNet} &
\textbf{ROCKET} &
\textbf{InceptionTime} &
\textbf{PDFTime(Ours)} \\
\midrule
MedicalImages & \textcolor{red}{0.9950} & 0.9750 & 0.9212 & 0.7511 & 0.9917 & 0.8677 & 0.7911 & 0.7950 & \textcolor{blue}{0.9923} \\
MelbournePedestrian & 0.7880 & 0.8823 & \textcolor{blue}{0.9532} & 0.8815 & 0.9213 & \textcolor{red}{0.9567} & 0.9031 & 0.9117 & 0.7617 \\
MiddlePhalanxOutlineAgeGroup & 0.6990 & 0.6573 & \textcolor{blue}{0.9170} & 0.6273 & \textcolor{red}{0.9573} & 0.8140 & 0.5532 & 0.5519 & 0.8822 \\
MiddlePhalanxOutlineCorrect & 0.7040 & 0.8092 & 0.9059 & 0.8474 & 0.8701 & \textcolor{blue}{0.9339} & 0.8289 & 0.8275 & \textcolor{red}{0.9986} \\
MiddlePhalanxTW & 0.8050 & \textcolor{blue}{0.9170} & 0.8377 & 0.5805 & \textcolor{red}{0.9204} & 0.7309 & 0.5403 & 0.5065 & 0.8752 \\
MixedShapesRegularTrain & 0.8830 & 0.7611 & \textcolor{blue}{0.9764} & 0.9494 & 0.8376 & 0.9053 & 0.9676 & 0.9563 & \textcolor{red}{0.9980} \\
MixedShapesSmallTrain & 0.8490 & 0.7318 & 0.9537 & 0.8934 & \textcolor{blue}{0.9757} & 0.9038 & 0.9347 & 0.9039 & \textcolor{red}{0.9858} \\
MoteStrain & 0.7440 & 0.6924 & \textcolor{red}{0.9843} & 0.8997 & \textcolor{blue}{0.9743} & 0.9623 & 0.9152 & 0.8778 & 0.8882 \\
NonInvasiveFetalECGThorax1 & 0.7540 & \textcolor{blue}{0.9757} & 0.9692 & 0.9539 & \textcolor{red}{0.9843} & 0.9349 & 0.9524 & 0.9376 & 0.8167 \\
NonInvasiveFetalECGThorax2 & 0.7440 & 0.9504 & 0.9679 & 0.9595 & \textcolor{blue}{0.9724} & 0.9575 & 0.9656 & 0.9532 & \textcolor{red}{1.0000} \\
OliveOil & \textcolor{red}{0.9790} & 0.9451 & 0.7333 & 0.9600 & \textcolor{blue}{0.9782} & 0.8500 & 0.9000 & 0.8333 & 0.4166 \\
OSULeaf & \textcolor{blue}{0.9690} & 0.6967 & 0.9526 & 0.8248 & 0.8333 & 0.8127 & 0.9347 & 0.9215 & \textcolor{red}{1.0000} \\
PhalangesOutlinesCorrect & 0.7530 & 0.8799 & 0.9267 & 0.8235 & \textcolor{blue}{0.9504} & 0.9316 & 0.8275 & 0.8399 & \textcolor{red}{1.0000} \\
Phoneme & 0.5440 & \textcolor{blue}{0.8619} & 0.5141 & 0.3387 & \textcolor{red}{0.9300} & 0.1981 & 0.2759 & 0.3204 & 0.7839 \\
PickupGestureWiimoteZ & 0.4670 & 0.2781 & \textcolor{blue}{0.8700} & 0.8160 & 0.7844 & 0.7200 & 0.8400 & 0.6160 & \textcolor{red}{1.0000} \\
PigAirwayPressure & 0.4800 & \textcolor{blue}{0.6300} & 0.5593 & 0.3750 & \textcolor{red}{0.9000} & 0.2446 & 0.3913 & 0.5529 & 0.6603 \\
PigArtPressure & 0.9830 & 0.3500 & 0.9458 & 0.9837 & 0.6432 & 0.3826 & 0.9519 & \textcolor{blue}{0.9952} & \textcolor{red}{1.0} \\
PigCVP & 0.8930 & 0.5308 & 0.8212 & 0.9029 & \textcolor{red}{0.9937} & 0.4181 & \textcolor{blue}{0.9327} & 0.9173 & 0.8959 \\
PLAID & \textcolor{red}{0.9670} & 0.5269 & 0.6369 & 0.9084 & \textcolor{blue}{0.9458} & 0.5951 & 0.8819 & 0.9128 & 0.6434 \\
Plane & 0.6370 & 0.4107 & \textcolor{blue}{0.9952} & \textcolor{red}{1.0000} & 0.6192 & 0.9810 & \textcolor{red}{1.0000} & \textcolor{red}{1.0000} & 0.8678 \\
PowerCons & 0.5080 & \textcolor{blue}{0.9952} & 0.9944 & 0.9467 & \textcolor{red}{1.0000} & 0.9722 & 0.9333 & 0.9444 & 0.8893 \\
ProximalPhalanxOutlineAgeGroup & 0.3460 & 0.9333 & 0.9190 & 0.8634 & \textcolor{blue}{0.9944} & 0.8860 & 0.8449 & 0.8107 & \textcolor{red}{1.0000} \\
ProximalPhalanxOutlineCorrect & 0.8780 & 0.8727 & \textcolor{blue}{0.9238} & 0.9003 & 0.9091 & \textcolor{red}{0.9406} & 0.9003 & 0.9107 & 0.9496 \\
ProximalPhalanxTW & 0.8420 & \textcolor{blue}{0.9170} & 0.9074 & 0.8117 & \textcolor{red}{0.9294} & 0.8512 & 0.7912 & 0.7366 & 0.8987 \\
RefrigerationDevices & \textcolor{red}{0.9940} & 0.8798 & 0.8333 & 0.5749 & 0.9273 & 0.5693 & 0.5141 & 0.5387 & \textcolor{blue}{0.9933} \\
Rock & \textcolor{red}{1.0000} & 0.6400 & 0.8286 & 0.8560 & 0.8027 & 0.8143 & \textcolor{blue}{0.9000} & 0.8200 & 0.8922 \\
ScreenType & \textcolor{red}{1.0000} & 0.6940 & 0.8267 & 0.5883 & 0.8286 & 0.6347 & 0.4677 & 0.5653 & \textcolor{blue}{0.9989} \\
SemgHandGenderCh2 & 0.9440 & 0.5093 & \textcolor{blue}{0.9744} & 0.8967 & 0.7027 & 0.9367 & 0.9223 & 0.8687 & \textcolor{red}{1.0000} \\
SemgHandMovementCh2 & \textcolor{red}{1.0000} & 0.9378 & 0.8578 & 0.6036 & 0.9778 & 0.7800 & 0.6298 & 0.4813 & \textcolor{blue}{0.9992} \\
SemgHandSubjectCh2 & \textcolor{red}{1.0000} & 0.6311 & 0.9556 & 0.8884 & 0.8256 & 0.9411 & 0.8676 & 0.7844 & \textcolor{blue}{0.9773} \\
ShakeGestureWiimoteZ & 0.9540 & 0.8278 & 0.9500 & 0.9040 & \textcolor{blue}{0.9633} & 0.5900 & 0.9000 & 0.8000 & \textcolor{red}{1.0000} \\
ShapeletSim & 0.9150 & 0.6900 & \textcolor{red}{1.0000} & \textcolor{red}{1.0000} & 0.9800 & 0.5500 & \textcolor{red}{1.0000} & \textcolor{blue}{0.9889} & \textcolor{red}{1.0000} \\
ShapesAll & 0.8840 & 0.9150 & 0.9083 & 0.8713 & \textcolor{red}{0.9950} & 0.8325 & 0.9033 & 0.9300 & \textcolor{blue}{0.9893} \\
SmallKitchenAppliances & 0.8000 & 0.8833 & 0.9567 & 0.7675 & \textcolor{blue}{0.9575} & 0.7840 & 0.8091 & 0.7611 & \textcolor{red}{1.0000} \\
SmoothSubspace & 0.5240 & 0.6387 & \textcolor{red}{0.9904} & 0.8880 & 0.9053 & 0.9767 & \textcolor{blue}{0.9773} & 0.9667 & 0.8759 \\
SonyAIBORobotSurface1 & 0.9620 & 0.9633 & \textcolor{blue}{0.9969} & 0.9464 & 0.9936 & 0.9952 & 0.9158 & 0.9334 & \textcolor{red}{0.9991} \\
SonyAIBORobotSurface2 & 0.8030 & 0.9888 & 0.9910 & 0.9324 & 0.9936 & \textcolor{blue}{0.9939} & 0.9196 & 0.9234 & \textcolor{red}{1.0} \\
StarLightCurves & 0.5510 & 0.9939 & 0.9817 & 0.9813 & \textcolor{blue}{0.9980} & 0.9616 & 0.9806 & 0.9777 & \textcolor{red}{1.0000} \\
Strawberry & 0.9670 & \textcolor{blue}{0.9912} & 0.9893 & 0.9757 & 0.9905 & 0.9766 & 0.9811 & 0.9768 & \textcolor{red}{1.0000} \\
SwedishLeaf & 0.6600 & 0.9624 & 0.9833 & 0.9533 & \textcolor{blue}{0.9868} & 0.9449 & 0.9616 & 0.9616 & \textcolor{red}{1.0000} \\
Symbols & 0.9520 & 0.9778 & \textcolor{blue}{0.9967} & 0.9574 & 0.9867 & 0.9618 & 0.9729 & 0.9739 & \textcolor{red}{1.0000} \\
SyntheticControl & 0.8970 & 0.9922 & 0.9778 & \textcolor{blue}{0.9993} & 0.9941 & 0.9767 & \textcolor{red}{1.0000} & 0.9980 & \textcolor{red}{1.0000} \\
ToeSegmentation1 & 0.9440 & 0.9700 & 0.9522 & 0.9518 & \textcolor{blue}{0.9967} & 0.8514 & 0.9518 & 0.9693 & \textcolor{red}{1.0000} \\
ToeSegmentation2 & 0.9670 & 0.9219 & \textcolor{red}{1.0000} & 0.9185 & \textcolor{blue}{0.9852} & 0.8316 & 0.9231 & 0.9308 & \textcolor{red}{1.0000} \\
Trace & \textcolor{red}{1.0000} & 0.8510 & \textcolor{blue}{0.9991} & \textcolor{red}{1.0000} & 0.9882 & 0.9400 & \textcolor{red}{1.0000} & \textcolor{red}{1.0000} & \textcolor{blue}{0.9991} \\
TwoLeadECG & 0.7985 & \textcolor{red}{1.0000} & \textcolor{blue}{0.9998} & 0.9991 & \textcolor{red}{1.0000} & 0.9983 & 0.9991 & 0.9965 & 0.8910 \\
TwoPatterns & 0.6528 & \textcolor{blue}{0.9991} & 0.9944 & 0.9961 & \textcolor{red}{1.0000} & 0.9980 & \textcolor{red}{1.0000} & \textcolor{red}{1.0000} & 0.8729 \\
UMD & 0.7239 & \textcolor{blue}{0.9990} & 0.9913 & 0.9944 & \textcolor{red}{1.0000} & 0.9833 & 0.9931 & 0.9653 & 0.8178 \\
UWaveGestureLibraryAll & 0.6201 & \textcolor{red}{1.0000} & 0.9413 & 0.9523 & 0.9944 & 0.9504 & 0.9739 & 0.9370 & \textcolor{blue}{0.9976} \\
UWaveGestureLibraryX & 0.7122 & \textcolor{red}{0.9913} & 0.9194 & 0.8211 & \textcolor{blue}{0.9900} & 0.8924 & 0.8557 & 0.8110 & 0.8829 \\
UWaveGestureLibraryY & 0.7810 & 0.9214 & 0.9219 & 0.7452 & \textcolor{blue}{0.9406} & 0.8455 & 0.7772 & 0.7499 & \textcolor{red}{0.9989} \\
UWaveGestureLibraryZ & 0.8654 & 0.8955 & \textcolor{red}{0.9999} & 0.7697 & \textcolor{blue}{0.9109} & 0.8669 & 0.7950 & 0.7552 & 0.8953 \\
Wafer & \textcolor{red}{1.0000} & 0.9006 & 0.7494 & 0.9978 & 0.9170 & \textcolor{blue}{0.9986} & 0.9979 & 0.9978 & 0.9264 \\
Wine & \textcolor{red}{1.0000} & \textcolor{blue}{0.9992} & 0.8928 & 0.8667 & \textcolor{red}{1.0000} & 0.8289 & 0.7815 & 0.7222 & 0.8630 \\
WordSynonyms & \textcolor{red}{0.9688} & 0.5135 & 0.8429 & 0.7213 & \textcolor{blue}{0.8842} & 0.8442 & 0.7467 & 0.7132 & \textcolor{red}{0.9688} \\
Worms & \textcolor{red}{1.0000} & 0.8690 & 0.8692 & 0.7896 & \textcolor{blue}{0.9175} & 0.7446 & 0.7273 & 0.7792 & \textcolor{red}{1.0000} \\
WormsTwoClass & \textcolor{red}{1.0000} & 0.8072 & \textcolor{blue}{0.9694} & 0.8286 & 0.8453 & 0.7833 & 0.8052 & 0.7792 & \textcolor{red}{1.0000} \\
Yoga & \textcolor{red}{1.0000} & 0.9037 & \textcolor{blue}{0.9694} & 0.8793 & 0.8802 & 0.9539 & 0.9066 & 0.9050 & \textcolor{red}{0.994} \\
\midrule
\textbf{Average Accuracy} & 0.8212 & 0.8255 & 0.9210 & 0.8508 & \textcolor{blue}{0.9326} & 0.8367 & 0.8450 & 0.8352 & \textcolor{red}{0.9461} \\
\textbf{Average Rank} & 5.8594 & 5.6641 & 3.4688 & 5.6094 & \textcolor{blue}{2.9609} & 5.8516 & 5.6250 & 6.1875 & \textcolor{red}{2.5938} \\
\textbf{Top-1 num} & 22.0 & 11.0 & 16.0 & 10.0 & \textcolor{blue}{26.0} & 6.0 & 11.0 & 7.0 & \textcolor{red}{80.0} \\
\bottomrule
\end{tabular}
}
\end{table*}

More details about the UEA and UCR datasets can be found in \url{https://www.timeseriesclassification.com/}.

\section{Fair Experimental Comparison}\label{app:fair}

To ensure a fair and reproducible comparison, we strictly follow the experimental
protocols and implementation settings adopted in prior works whenever possible.
For baseline methods whose official implementations and hyperparameter configurations are publicly available, we directly use the released GitHub repositories and the parameter settings recommended by the original authors. Specifically, we use the deep learning framework provided in their official GitHub \url{https://github.com/thuml/Time-Series-Library} repository, where the training and validation set splits are also derived from.

For models provided in the Time-Series-Library framework
\cite{timesnet}, we adopt the original training pipelines, optimizer settings,
and data preprocessing strategies without modification.
For methods whose results are reported in large-scale benchmark studies
\cite{ma2024survey}, we directly use the published test accuracy results
to avoid discrepancies introduced by reimplementation details. To ensure the reproducibility of our results, all experiments for PDFTime are conducted using a fixed random seed of 2025.

For baseline models without publicly released hyperparameter configurations,
we employ the default settings suggested in the corresponding papers or
official codebases. No dataset-specific tuning is performed for any baseline method.

For the proposed PDFTime framework, we adopt a unified set of hyperparameters
across all datasets, as summarized in Table~\ref{tab:exp_settings}.
This configuration is kept fixed throughout all experiments and is not tuned
for individual datasets.

All models are trained and evaluated under identical data splits and evaluation
metrics. This protocol ensures that the reported performance differences are
attributable to modeling choices rather than implementation details or
hyperparameter tuning advantages. 

Specifically, the experiments on the UEA benchmark were conducted on NVIDIA Tesla V100 (32GB) GPUs, while the experiments on the UCR archive were performed on NVIDIA A100 (80GB) GPUs. Despite the difference in peak floating-point performance between these platforms, we ensure that the training convergence and final accuracy are consistent across devices.
\begin{table}[!htbp]
\centering
\caption{Unified hyperparameter configuration used for the proposed PDFTime framework across all datasets.}
\label{tab:exp_settings}
\begin{tabular}{lc}
\toprule
\textbf{Hyperparameter} & \textbf{Value} \\
\midrule
Number of encoder layers ($L_e$) & 2 \\
Batch size & 16 \\
Dropout rate & 0.2 \\
Model dimension ($d_{\text{model}}$) & 128 \\
Learning rate & 0.001 \\
Maximum training epochs & 150 \\
Early stopping patience & 20 \\
\bottomrule
\end{tabular}
\end{table}

\section{Analysis of Prototype Momentum Hyperparameters}\label{app:prototype}

This section analyzes the key hyperparameters governing the proposed prototype momentum update strategy.
These hyperparameters control the temporal dynamics of prototype adaptation and are conceptually related to commonly used mechanisms such as exponential moving averages and learning rate schedules, which facilitates interpretation and practical tuning.

\subsection{Prototype Hierarchy and Cardinality}

Beyond the momentum schedule itself, the structure of the prototype hierarchy and the number of prototypes per class play an important role in determining model behavior.
In our framework, prototypes are organized in a shallow hierarchical manner.
Lower-level prototypes are designed to capture fine-grained intra-class variations, while higher-level prototypes represent more abstract and stable class-level patterns.

In the main experiments, we adopt a two-layer prototype hierarchy.
For a $C$-class classification task, the first layer contains $2 \times C$ prototypes, and the second layer contains $3 \times C$ prototypes.
This design follows a coarse-to-fine abstraction principle: the lower layer emphasizes diversity and local clustering, whereas the upper layer enforces stronger class-level consistency through hierarchical constraints.

We do not extensively explore deeper hierarchies or substantially larger prototype sets, as such configurations significantly increase computational cost and introduce additional complexity in hierarchical coordination.
Moreover, we observe that the appropriate hierarchy depth and prototype cardinality are closely related to dataset characteristics, including dataset scale, noise level, and intra-class feature dispersion.
Nevertheless, we evaluate several alternative prototype configurations to assess robustness, and the corresponding results are summarized in Table~\ref{tab:app_gamma}.
Overall, the two-layer design provides a favorable balance between representation capacity and training stability across representative UEA datasets.

\subsection{Dynamic Momentum Scheduling}

The prototype update process is controlled by a time-dependent momentum coefficient.
The scheduling variable $t$ reflects training progress and can be defined at either the epoch or iteration level.
Accordingly, the training process is divided into a warm-up phase and an active phase, analogous to warm-up strategies commonly used in learning rate scheduling.

During the warm-up phase, prototype updates are frozen to prevent unstable early representations from dominating prototype states.
Once the active phase begins, prototypes are allowed to adapt more rapidly to emerging feature clusters.
The momentum coefficient $\gamma$ determines the update rate of the exponential moving average.
Smaller values of $\gamma$ encourage faster adaptation, while larger values emphasize stability and long-term memory.

The lower bound $\gamma_a$ specifies the maximum responsiveness during the active phase, whereas the upper bound $\gamma_b$ controls stabilization in later training stages.
The transition between these regimes is regulated by a time constant $\tau$, which enables a smooth and gradual change in prototype dynamics rather than abrupt switching.

\subsection{Sensitivity Analysis and Robustness}
\begin{table}
\centering
\caption{Effect of prototype update strategy ($\gamma$).
Results are averaged over 10 UEA datasets.}
\label{tab:app_gamma}
\begin{tabular}{lc}
\toprule
\textbf{Update Strategy} 
& \textbf{Avg. Acc (\%)}  \\
\midrule
$\gamma_a = 0.7, \gamma_b = 0.99$  & 0.755 \\
$\gamma_a = 0.95, \gamma_b = 0.999$  & 0.766 \\
$K^{(l)}=\{3,3,3\}$ &  0.763 \\
$K^{(l)}=\{5,3\}$ &  0.772 \\
$K^{(l)}=\{3,3\}$ &  0.780 \\
$K^{(l)}=\{2,2\}$ &  0.778 \\
\midrule
\textbf{Adaptive $\gamma$ and $K^{(l)}=\{2,3\}$(PDFTime)} & \textbf{0.783} \\
\bottomrule
\end{tabular}
\end{table}

We analyze the sensitivity of the proposed momentum schedule with respect to its key hyperparameters, including $\gamma_a$, $\gamma_b$, $T_{\text{warm}}$, $T_{\text{active}}$, and $\tau$.
The goal of this analysis is to evaluate robustness rather than to identify dataset-specific optimal configurations.

Experiments are conducted on a subset of representative datasets from the UEA multivariate time series classification benchmark.
In each experiment, one hyperparameter is varied while the others are fixed.
Table~\ref{tab:app_gamma} reports the resulting classification accuracy under different settings.

In addition to momentum-related parameters, we also examine the influence of the prototype hierarchy configuration.
Let $L_p$ denote the number of prototype layers, and let $K^{(l)}$ represent the number of prototypes per class at the $l$-th layer, where $l \in \{1, \dots, L_p\}$.
For a $C$-class classification task, the total number of prototypes is given by
\begin{equation}
N_p = \sum_{l=1}^{L_p} K^{(l)} \cdot C .
\end{equation}
In the main configuration, we adopt a two-layer hierarchy ($L_p = 2$) with $(K^{(1)}, K^{(2)}) = (2, 3)$, resulting in $N_p = 5C$ prototypes in total.

The results indicate that model performance is generally stable under moderate variations of $\gamma_a$ and $\gamma_b$.
Specifically, accuracy remains comparable when $\gamma_a$ varies within $[0.90, 0.99]$ and $\gamma_b$ within $[0.995, 0.999]$.
Similar trends are observed for $T_{\text{warm}}$ and $T_{\text{active}}$, provided that a short warm-up phase is retained.

The hyperparameter values used in the main experiments (i.e., $T_{\text{warm}} = 3$, $T_{\text{active}} = 10$, $\gamma_a = 0.97$, $\gamma_b = 0.997$, $\tau = 30$, and $(L_p, K^{(l)}) = (2, \{2,3\})$) are intended as a reference configuration for UEA-style datasets.
We note that alternative datasets may benefit from different settings depending on factors such as signal noise, temporal structure, and intra-class variability.

\section{Additional Prototype interpretability and Analysis}\label{app:visual}

To further support the observations made in the main paper, we provide additional qualitative analyses of the learned prototypes and embedding space. These visualizations aim to offer deeper insights into how prototypes interact with data distributions and how diversity is maintained within and across classes.

\subsection{Extended figure for PDFTime interpretability}
\begin{figure}[t]
    \centering
    \includegraphics[width=0.8\linewidth]{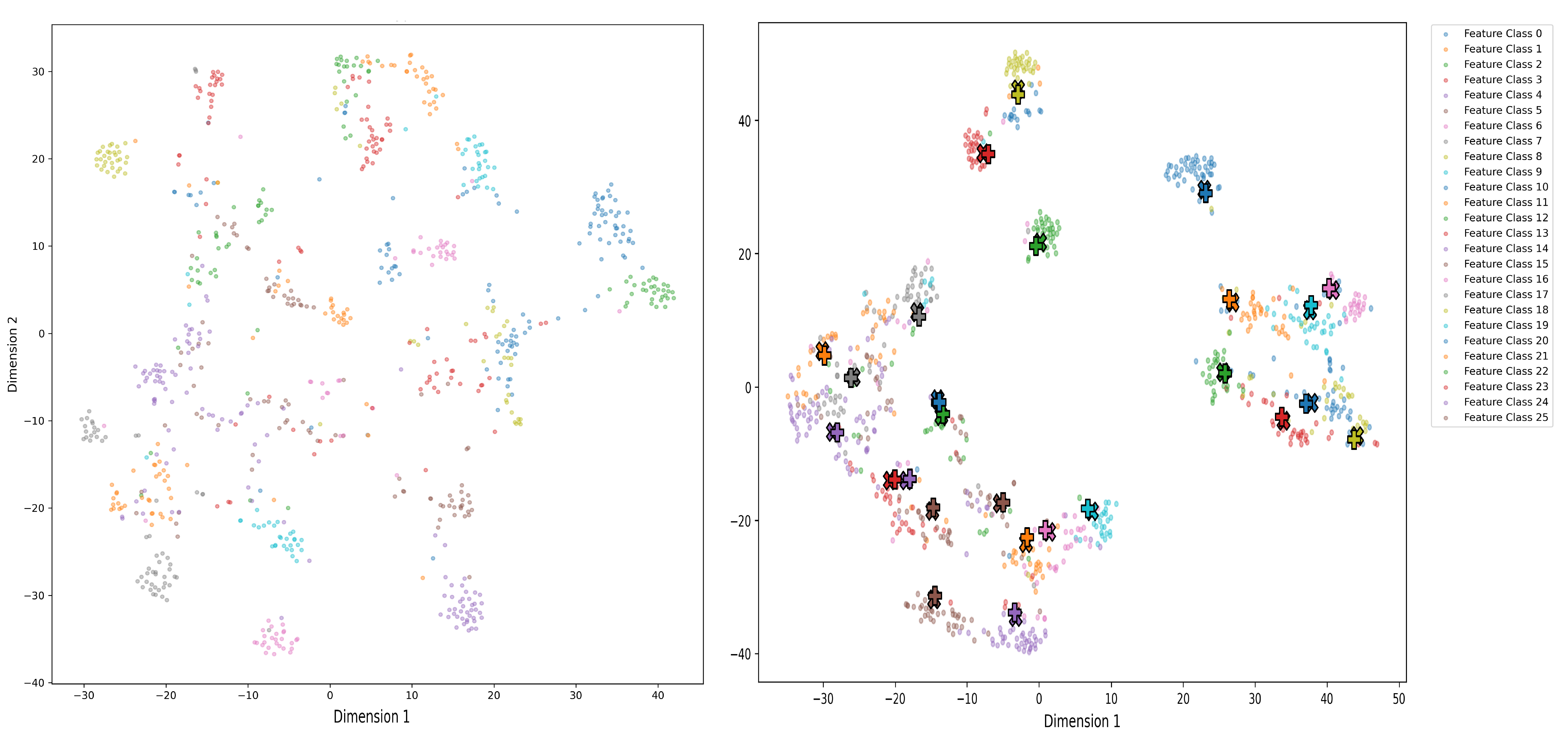}
    \caption{t-sne on dataset Handwriting.The left figure is from the MLP classification head, and the right figure is ours. The prototype is represented by X in the diagram.}
    \label{fig:appendix-tsne}
\end{figure}
\begin{figure}
    \centering
    \includegraphics[width=0.8\linewidth]{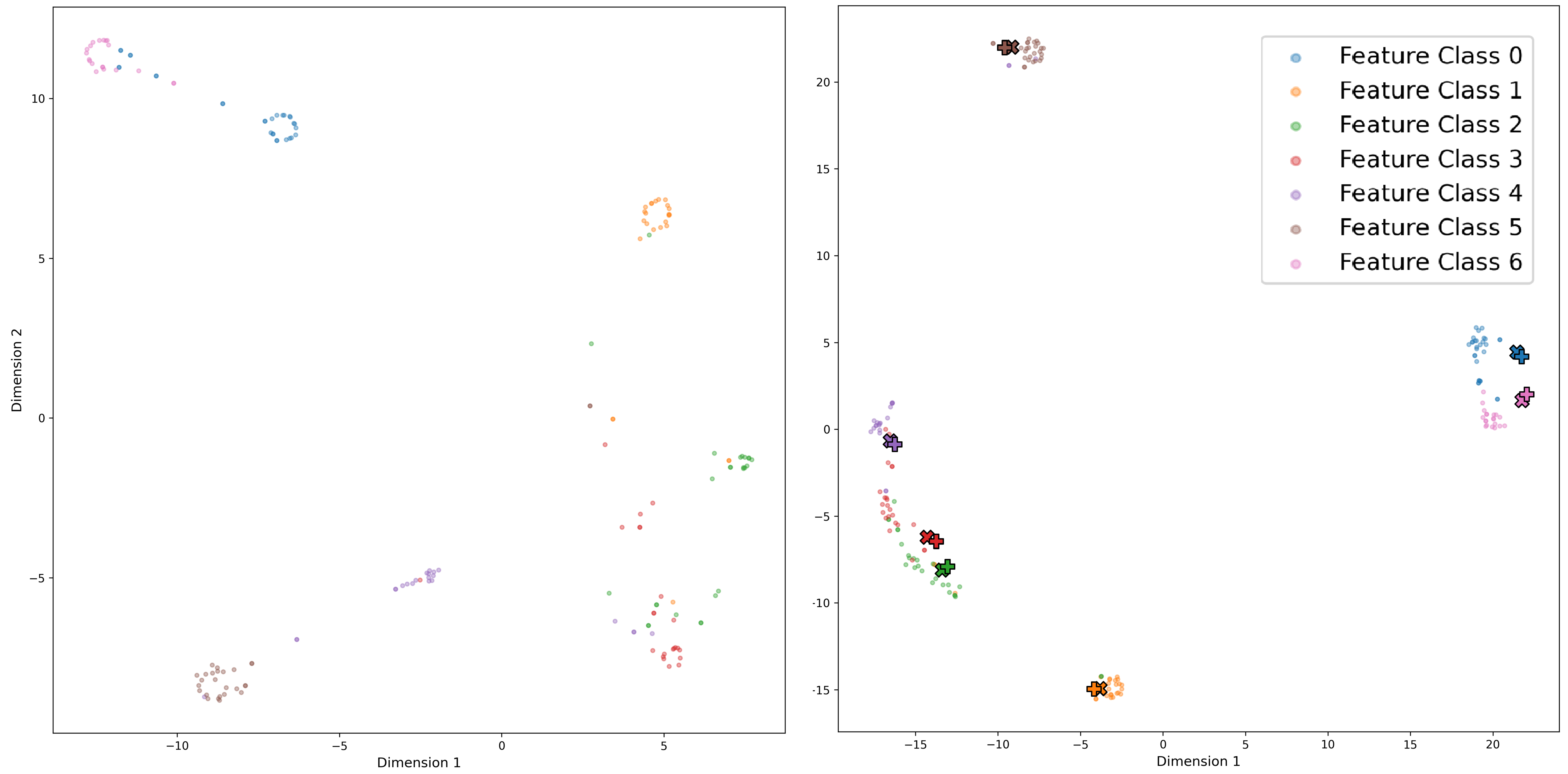}
    \caption{t-sne on dataset PSMMS-SF.The left figure is from the MLP classification head, and the right figure is ours. The prototype is represented by X in the diagram.}
    \label{fig:appendix-tsne2}
\end{figure} 
\begin{figure}
    \centering
    \includegraphics[width=0.8\linewidth]{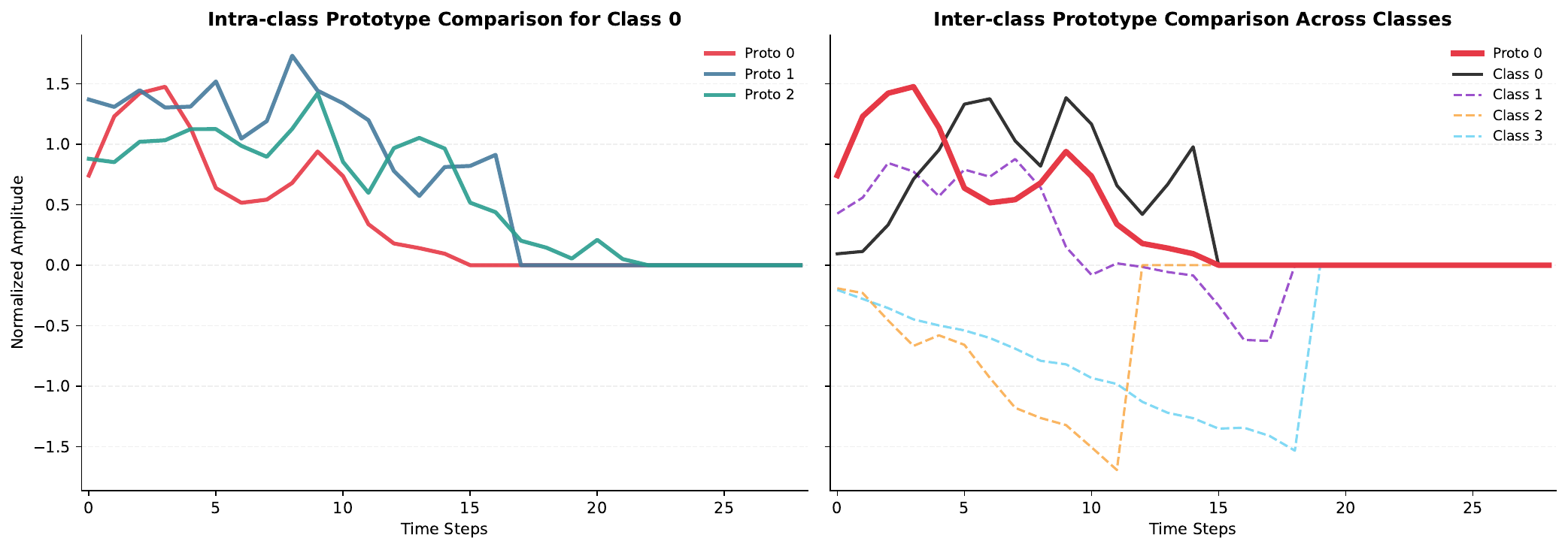}
    \caption{Comparison between learned prototypes and original time series on dataset JapaneseVowels.}
    \label{fig:cp1}
\end{figure}
\begin{figure}
    \centering
    \includegraphics[width=0.8\linewidth]{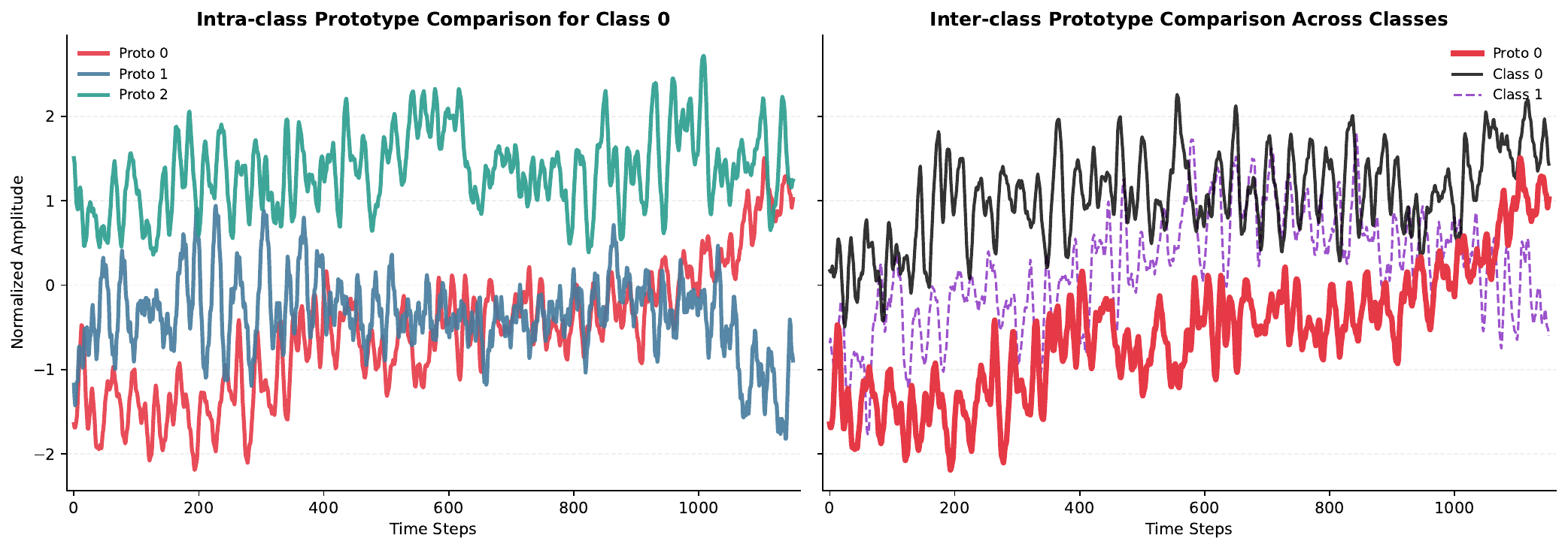}
    \caption{Comparison between learned prototypes and original time series on dataset SelfRegulationSCP2.}
    \label{fig:cp2}
\end{figure}
\begin{figure}
    \centering
    \includegraphics[width=0.8\linewidth]{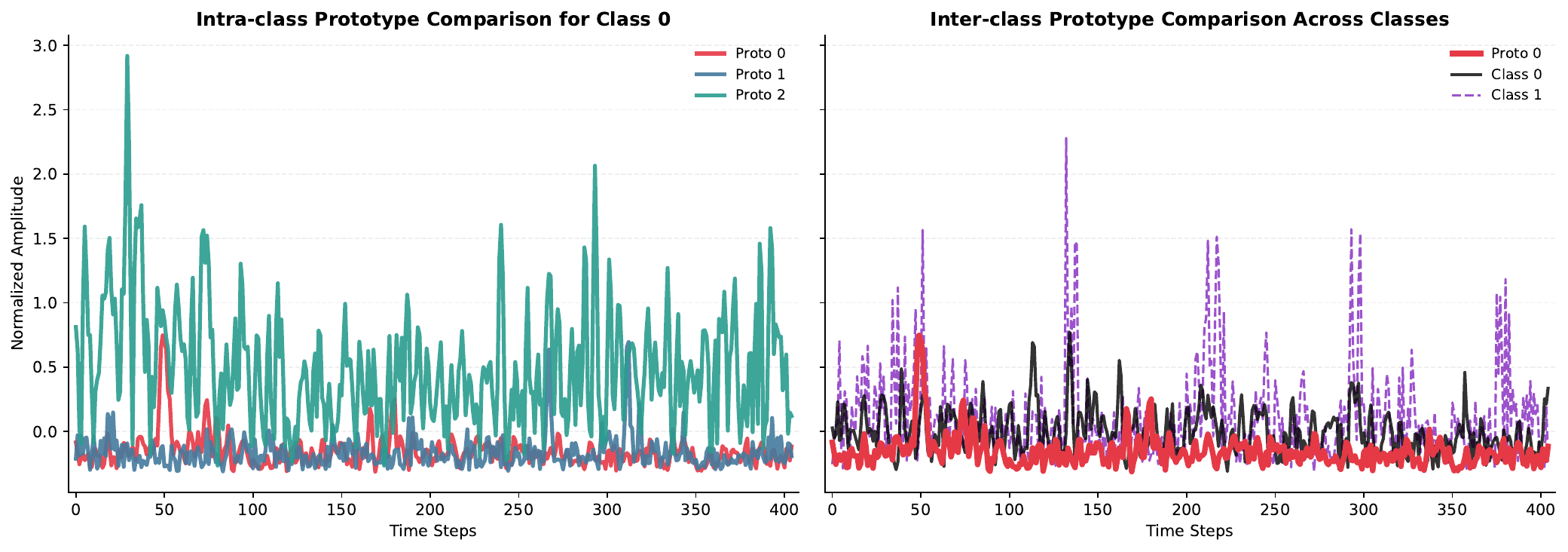}
    \caption{Comparison between learned prototypes and original time series on dataset Heartbeat.}
    \label{fig:cp3}
\end{figure}
Fig.~\ref{fig:appendix-tsne}-\ref{fig:appendix-tsne2} and Fig.~\ref{fig:cp1}-\ref{fig:cp3} present additional visualizations of the learned embeddings and their associated prototypes across different datasets, including both T-SNE projections and direct comparisons between prototypes and original time-series samples.
For the T-SNE visualizations, the settings are identical to those used in the main paper: the left figure shows embeddings obtained from the MLP classification head, while the right figure corresponds to our prototype-based head.
For the comparison plots, the left figure illustrates differences among prototypes within the same class, whereas the right figure compares a single prototype with all corresponding original time-series samples.
Across all settings, we observe consistent structural patterns.

\subsection{Dynamic Evolution of Prototypes} To investigate the convergence behavior of our framework, we visualize the joint evolution of feature embeddings and prototypes during the training process. Fig. \ref{fig:epoch40}-\ref{fig:epoch120} captures the state of the latent space at every 20 epochs.

Initially (Epoch 40), prototypes are often clustered near the center of the latent space with high overlap, representing an unrefined global mean. As training progresses (Epochs 60--100), the prototypes progressively "migrate" toward high-density regions of their respective class-conditional distributions. By the final stage (Epoch 100+), the prototypes successfully stabilize at the semantic centroids of distinct clusters, effectively capturing representative temporal motifs. This transition from stochastic chaos to structured order demonstrates the effectiveness of our EMA-based update strategy in distilling stable class-specific knowledge.

    
    
    
    
    

\subsection{Stability and Consistency Across Runs}

To examine the robustness of the learned prototype structure, we visualize multiple training runs with different random initializations. Although the absolute positions of the embeddings vary due to the stochastic nature of T-SNE, the relative relationships remain stable: prototypes consistently align with their respective class clusters and preserve both intra-class diversity and inter-class separation. This consistency indicates that the observed behaviors are intrinsic properties of the model rather than artifacts of a specific run.

\subsection{Implications for Interpretability}

These additional visual results further reinforce the interpretability of our approach. Each prediction can be associated with a small set of nearby prototypes, which correspond to representative patterns within the data distribution. Compared to post-hoc explanation methods, this prototype-based transparency is intrinsic to the model and persists across datasets and training runs, making the decision process more reliable and easier to interpret.

\section{Limitations and Future Work}

Despite its effectiveness, PDFTime has several limitations that suggest directions for future work.
First, the number of prototypes at each granularity level is manually specified. Although the model shows stable performance across reasonable settings, developing adaptive prototype allocation strategies could further improve flexibility.
Second, while PDFTime enhances interpretability at the feature-to-decision stage through prototype similarity, the temporal feature extraction process itself remains implicit. Combining the proposed framework with complementary temporal explanation techniques is a potential extension.
Finally, this work focuses on offline time series classification under subject-independent settings. Extending PDFTime to other scenarios, such as online or continual learning, is an interesting direction for future research.

\end{document}